\documentclass{article}
\usepackage{amsmath,graphicx,color}
 \DeclareGraphicsExtensions{.pdf,.jpeg,.png}
 \usepackage[multidot]{grffile}
 \usepackage{color}

\def\f{{\mathbf f}}

\def\etal{\text{\it et al.~}}

\title{CFA Bayer image sequence denoising and demosaicking chain}
\date{}
\author{A. Buades \and J. Duran}

\begin{document}
%
\maketitle
\begin{abstract}
 The demosaicking provokes the spatial and color correlation of noise, which is afterwards enhanced by the imaging pipeline.  The correct removal previous or simultaneously with the demosaicking process is not usually considered in the literature.  We present a novel imaging chain including a denoising of the Bayer CFA and a demosaicking method for image sequences. The proposed algorithm uses a spatio-temporal patch method for the noise removal and demosaicking of the CFA.  The experimentation, including real examples, illustrates the superior performance of the proposed chain, avoiding the creation of artifacts and colored spots in the final image.
\end{abstract}

\graphicspath{{./figures/}}

\section{Introduction} \label{sec:intro}

Digital photographs are usually represented by three color values at each pixel. However, most common cameras use a CCD or CMOS sensor device measuring a single color per pixel. Demosaicking is the interpolation process by which the two missing color values are estimated. The selected configuration of the sensor usually follows the Bayer color filter array (CFA) \cite{Bayer}. Thus, out of a group of four pixels, two are green (in quincunx), one is red and one is blue \cite{AlleyssonSusstrunkHerault}. 

The demosaicking is usually performed by combining close values from the same channel or the other two. As a result, the noise, being almost white at the sensor, gets correlated.  The rest of the imaging chain, consisting mainly in color and gamma corrections and compression, enhances the noise in dark parts of the image leading to contrasted colored spots of several pixels. The size of these spots depends on the applied demosaicking method. The removal of these artifacts after the imaging chain is applied is a difficult task, since structure and noise are not easily differentiated. As a consequence, details and texture might be  attenuated during the denoising process, when having access only to this processed data.

In dark and indoor scenes, the limitations of imaging pipelines are more noticeable.  If the camera is set to a long exposure time, the photograph gets
blurred by the camera motion and aperture. If it is taken with short exposure, the image is dark, and enhancing it reveals the noise. The use of a high ISO value for short exposures makes the image brighter but is makes noise more annoying.  Many times, the dilema can be solved by taking a burst of images. A ``burst'', or ``image burst'' means a set of digital images taken from the same
camera, in the same state, and quasi instantaneously.  Such bursts are obtained by video, or by using the burst mode proposed in recent reflex and compact cameras, or simply by taking several snapshots while holding firmly the camera in a fixed direction.

We propose a multi-image processing chain, consisting of a novel denoising method for the removal of noise at the CFA sensor data and a novel multi-image demosaicking algorithm.  The denoising method deals automatically with the incomplete CFA and unknown signal dependent noise distribution.  It is based on the white noise removal algorithm for video \cite{vdenoisingTIP15}.  This algorithm groups similar spatio-temporal patches and removes noise by thresholding in an adapted PCA basis.

The novel demoisaicking method is also patch based. It combines similar patches by averaging, introducing the CFA Bayer pattern as a restriction to ensure that only initial values are combined.    The demosaicking process interpolates missing values but also modifies CFA ones. This permits the removal of residual noise and  avoids the creation of zipper effect.  The zipper effect is an on-off pattern created by the juxtaposition of contiguous original and interpolated pixels.   The proposed process is applied first to the green channel of the sequence, and then to the difference between red/blue channels and the updated green.

This paper is organized as follows. In Section \ref{sec:previous} we present the literature on image sequence and video denoising.  The new algorithm for denoising a sequence of color filter array (CFA) images is introduced in Section \ref{sec:cfadenoising}.  Section \ref{sec:demosaicking} introduce the novel image sequence demoisaicking method.
Section \ref{sec:experiments}  shows the performance of the proposed method by means of a comparison with state of the art approaches. 
Finally, Section \ref{sec:conclusions} concludes the paper.

\section{State of the Art} \label{sec:previous}

\subsection{Image demosaicking}

Color demosaicking techniques have been extensively reviewed in the literature \cite{MenonCalvagno, LiBai2017}. Due to the CFA configuration, the interpolation of the green channel is commonly performed in a first step. Then, in order to take advantage of inter-channel correlation, the differences or ratios between the red/blue and the estimated green are interpolated instead of the red/blue channel directly. Former demosaicking methods \cite{HamiltonAdams} concentrated on locally estimating the most suitable direction of interpolation for the green channel. In many cases, the ambiguity of the local configuration in the CFA makes nearly impossible to decide properly. Many authors \cite{ZhangWu, MenonAndrianiCalvagno} proposed to take this decision a posteriori once the green channel or even the full color image has been interpolated completely horizontally and completely vertically. 

Local demosaicking strategies can create artifacts such as aliasing, erroneously interpolated structures or \textit{zipper effect}. Buades \etal\cite{BuadesCollMorelSbert} showed that these interpolation artifacts can be eliminated by involving image self-similarity and redundancy. In a posterior paper \cite{duran2014self}, it was shown that nonlocal filtering achieves better results if applied to channel differences instead of channels themselves.

\subsection{Image sequence demosaicking}

Despite the extensive literature in single color image demosaicking, there exist few works on its extension to video. While the demosaicking of a single image might give reasonable results, the consecutive play of several frames might introduce visual artifacts. Wu \etal\cite{wu2006temporal} match the CFA green sample blocks in adjacent frames via motion analysis and fuse them with intra-frame estimates of the missing green samples. Lukac \etal\cite{lukac2006adaptive} introduced a video demosaicking method that uses a set of stencils comprising three consecutive frames. Gevrekci \etal\cite{gevrekci2007pocs} proposed an image sequence extension of the projection onto convex sets algorithm \cite{GunturkAltunbasakMersereau} by adding a new constraint set based on the spatio-intensity neighborhood. Vandewalle \etal  \cite{vandewalle2007joint} aligns the set of image looking for a rotation plus translation in the fourier domain.  After image registration, a demosaicked image is reconstructed at once using the full set of images. For this, the authors use normalized convolution, an image interpolation method from a nonuniform set of samples.  

Although it is beyond the scope of this paper, let us mention that video demosaicking is often proposed conjointly with super-resolution \cite{gotoh2004direct, farsiu2006multiframe, karch2015robust}. Gotoh \etal\cite{gotoh2004direct} introduced a variational method accounting for motion, subsampling and mosaicking with edge-adaptive regularization for luminance component and isotropic regularization for chrominance components. Farsiu \etal\cite{farsiu2006multiframe} proposed a hybrid method for super-resolution and demosaicking based, with the limitation of the motion field to be translational, on a MAP estimation technique by minimizing a cost function with bilateral regularization on the luminance component and Tikhonov regularization on the chrominance ones. An extra regularization term is incorporated to force similar edge location and orientation in different color channels.
Karch \etal\cite{karch2015robust} first demosaick all frames individually and interpolate all of them to the desired resolution. Then, a weighted sum of the interpolated frames is used to fuse them into an improved resolution estimate. Finally, restoration is applied to improve any degrading camera effects.

\subsection{Image sequence white noise removal}

Local average methods, as the bilateral filter \cite{tomasi1998bilateral},  or patch based methods as NL-means  \cite{NLmeans} or  BM3D \cite{dabov2009bm3d} and NLBayes \cite{lebrun2013nonlocal} can be easily adapted to video just by extending the neighboring area to the adjacent frames.    Kervrann and Boulanger \cite{boulanger2007space} extended the NL-means to video by growing adaptively the spatio-temporal neighborhood. Arias et al. extended the NL-Bayes \cite{lebrun2013nonlocal} to video \cite{arias2015towards, arias2017video}. Methods based in sparse decompositions are extended to image sequences  \cite{mairal2008learning, protter2009image, wen2015video, lee2015color}, as well as approaches based on low rank approximation \cite{ji2011robust, ji2010robust}.

Other methods combine a single image estimate with a purely temporal one.  Dai et al. \cite{dai2013generalized}  applied a purely temporal LMMSE estimate on motion trajectories.   Yue et al.  \cite{yue2015image} used a BM3D estimate of each frame and a  BM3D applied to similar patches of neighboring frames.
Dai et al. \cite{dai2013color} adds intercolor prediction to previous approaches.

The performance of many denoising methods is improved by  introducing motion compensation.   These compensated filters estimate explicitly the motion of the sequence and compensate the neighborhoods yielding stationary data \cite{ozkan1993adaptive}.  The BM3D extension, VBM4D \cite{maggioni2011video},  exploits the mutual similarity between 3-D spatio-temporal volumes constructed by tracking blocks along trajectories defined by the motion vectors. 
In \cite{vdenoisingTIP15} the authors proposed to combine optical flow estimation and patch based methods for denoising. 
Their algorithm tends to a fusion of the neighboring frames in the absence of occlusions and a dense temporal sampling.  In this ideal scenario, an optical flow or global registration is able to align the frames and fusion is achieved by simple averaging \cite{haro2012photographing}.  The algorithm in \cite{vdenoisingTIP15} compensates the failure of these requirements by introducing spatiotemporal patch comparison and denoising in an adapted PCA based transform.

\subsection{Image sequence color and spatially correlated noise removal}

It must be noted that the previous techniques apply only to additive uniform white noise, and not to real photography and video data. The literature on noise removal from  video is scarce. Bennet and McMillan \cite{bennett2005video} apply a spatiotemporal bilateral filtering combined with Durand et al. \cite{durand2002fast} enhancement technique. 
Filtering parameters at each pixel are fixed depending on the amount of enhancement to be applied.    Liu et al.  \cite{liu2010high} proposed a strategy using motion compensation and the NL-means algorithm.  Xu et al. \cite{xu2010new}  apply the NL-means denoising algorithm to each image  and then combine this estimate with a purely temporal application of the same algorithm. Both estimates are combined in static parts of the image, while the spatial estimate is preferred in moving ones.  A second iteration of the denoising algorithm is applied after a contrast enhancement stage.  A similar approach is used by Kim et al. \cite{kim2015novel} but a purely temporal denoising algorithm is applied first,  and a second purely spatial after the enhancement stage. Gao et al. \cite{gao2015video} perform a bilateral filter in both luminance and chrominance (YCbCr)  for each frame, for which the bilateral weight distribution is computed only on the luminance. A multiscale wavelet transform permits to deal with non white noise.
Jovanov et al. \cite{jovanov2015multiview} simultaneously perform multiview image sequence denoising, color correction and the improvement of sharpness in slightly defocused regions for sequences of images coming from different cameras and thus with different degrees of blur and noise. In \cite{buades2017denoising} the authors proposed a multi-scale algorithm, estimating and removing noise at each scale. It uses a variance stabilization transform and the white noise removal algorithm in \cite{vdenoisingTIP15}. 

A particular type of algorithms are those dealing with a burst of images. That is a sequence of images taken nearly instantaneously with close optical centers.  For such type of images, a parametric transform, commonly an homography, is able to register the images of the sequence.  A robust combination of these images after registration permits to remove correlated noise \cite{haro2012photographing}.  A fast version by  Liu et al. \cite{liu2014fast} accelerates the homography selection by a multi scale strategy. The multiscale procedure includes a local refining of the homography and the image combination.

\subsection{Noise removal at the CFA and joint denoising-demosaicking}

Since demosaicking is the main cause of noise correlation, it is suitable to remove the noise before this process, or simply combine them in a single procedure. 

Paliy et al. \cite{paliy2008denoising} performs interpolation and denoising of a single CFA Bayer image.  The method uses inter color filters selected by polynomial approximation.   Chatterjee et al. \cite{chatterjee2011noise}  denoises the CFA by using the NL-means algorithm. Their method averages only patches  having the same CFA pattern. The authors do not use any variance stabilization transform.  They propose a demosaicking method based on an optimization process where the CFA is taken as the low resolution counterpart in a super resolution framework.

Zhang \etal\cite{zhang2010spatial} proposed to first denoise the CFA before applying a spatio-temporal demosaicking algorithm. A spatio temporal extension of \cite{zhang2010two} is applied for denoising by combining only patches having the same CFA pattern. Noise is reduced by thresholding in an adaptive PCA basis. Since each patch belongs to the CFA actually contains red, green and blue pixels.  For demosaicking, an initial demosaicking is processed by a spatial-temporal algorithm in order to reduce artifacts.  For a given pixel to be processed,  one searches for the similar pixels to it within the spatial-temporal neighborhood and then let the enhanced pixel be the weighted average of them.

Heide et al. \cite{heide2014flexisp}  proposed a single optimization process being able to deal with all image chain stages.  Denoising, demosaicking and deconvolution are written in a single energy minimization solved by primal-dual techniques \cite{chambolle2011first}. The method applies to a single image, even it can be adapted to deal with image sequences. Patil et al. \cite{patil2016poisson} denoise the CFA sensor data by a dictionary learning combined with a variance stabilization transform.  No particular demosaicking algorithm is proposed.
Hasinoff et al. \cite{hasinoff2016burst} proposed a chain for burst sequences of images, with the novelty of dealing with the Bayer data. The method aligns and merges all images of the burst sequence into a single one.  Subsampled Bayer of factor 2 is used for alignment, a tiled translation is estimated by a Gaussian pyramid process.  Finally, a robust merging method is implemented using the FFT.  

Neural networks have also been recently proposed for burst fusion Mildenhall et al. \cite{mildenhall2017burst}.

\section{CFA Denoising}\label{sec:cfadenoising}

We propose to denoise the sequence of CFA images  before the demosaicking.  In order to do so, instead of dealing directly with the CFA structure, we subsample each image into a 4-channel one, accounting for the red, two greens and blue values. Each new image has half the width and height of the original sensor data.  Now, a sequence of complete 4-channel images has to be denoised.

\begin{figure} [t]
\centering
\includegraphics[width=4 cm]{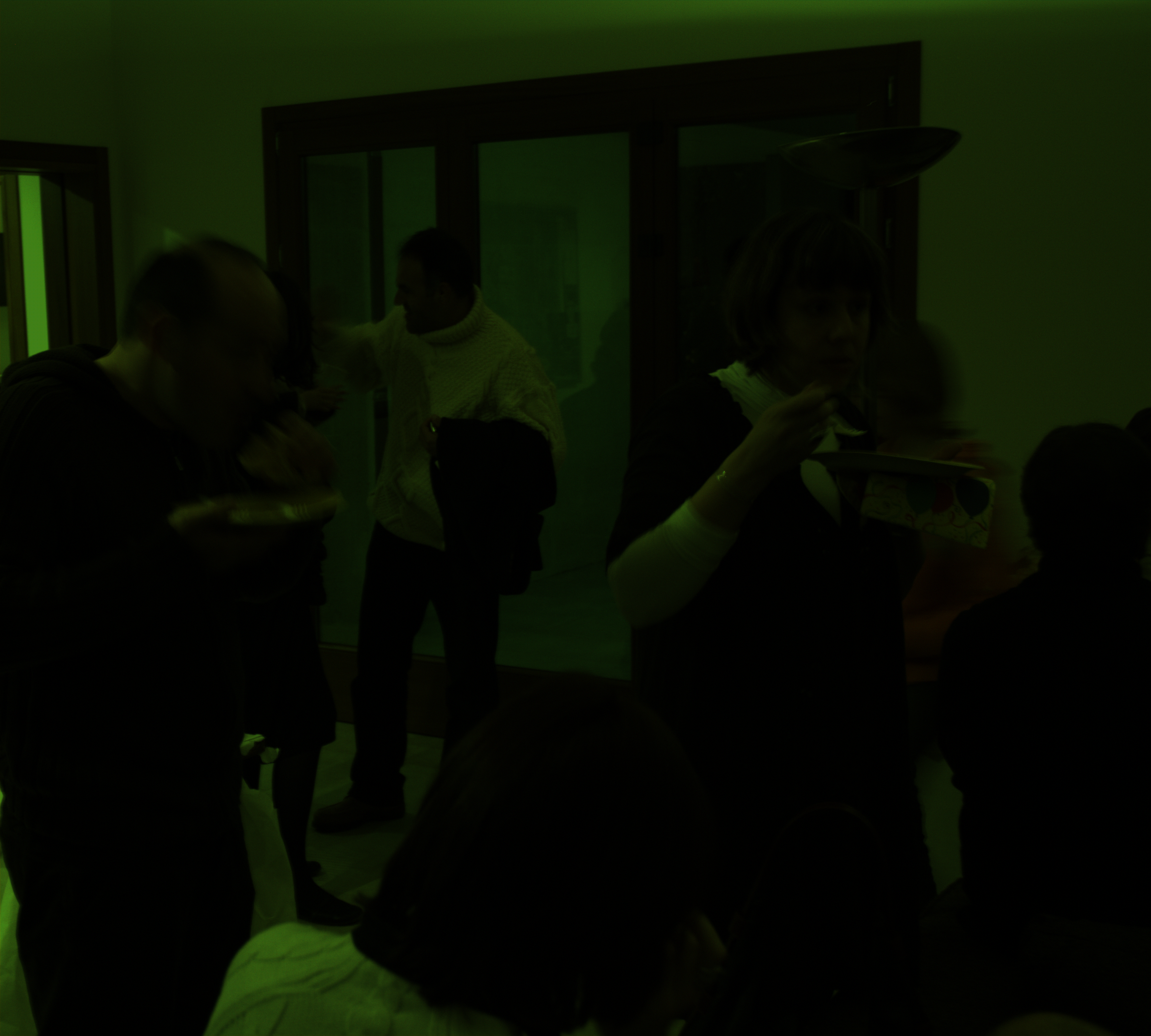}

\includegraphics[width=4 cm]{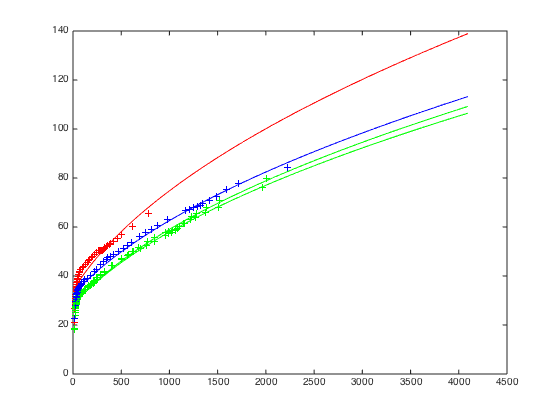}
\includegraphics[width=4 cm]{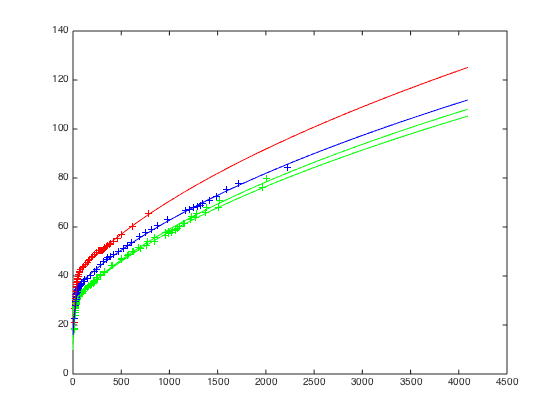}

\includegraphics[width=4 cm]{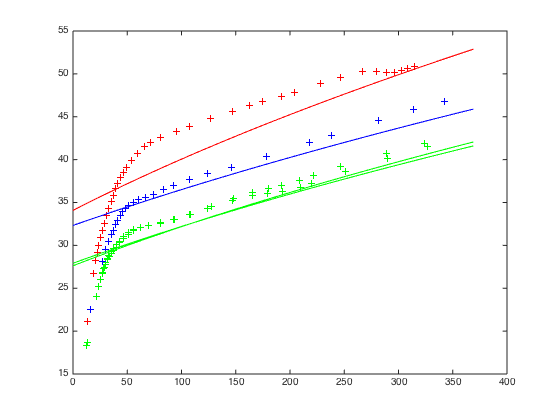}
\includegraphics[width=4 cm]{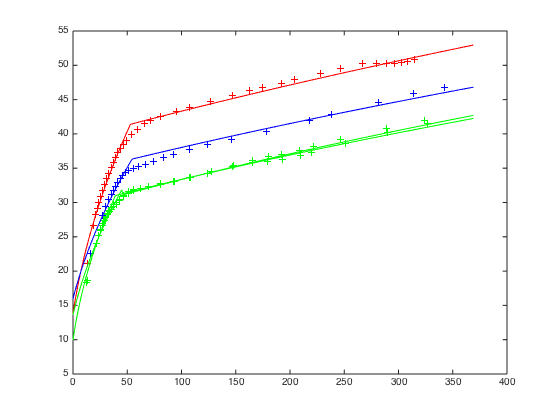}
\caption{Noise estimation for a RAW image. Top: image. Below: single linear variance model and proposed approach. Bottom: zoom on the low part of the range. It is clear that a single function is not able to correctly approximate the whole range. This single approximation does not fit correctly the darker part, which is the most important for the denoising stage.} \label{fig:noiseint}
\end{figure}

\subsection{Noise estimation}
 
 In order to estimate noise, we assume that all images of the sequence have been acquired with the same conditions, that is, the same ISO gain factor and exposure. In such a case, all images of the sequence share the same noise model. 

We adapt the single image noise estimation algorithm in \cite{colom2014nonparametric}. This algorithm divides each image in patches and applies the DCT as proposed by Ponomarenko \cite{ponomarenkoLZKA07} for uniform noise estimation.  The low frequencies of the DCT permit to select the less oscillating patches, and the high frequencies of these selected patches yield a standard deviation estimate.

We classify patches from all the sequence depending on its mean.  The full color range is partitioned into non overlapping bins, and each patch is arranged into the corresponding one. This permits to estimate a standard deviation for each bin, and thus an intensity dependent model.   Since noise at the sensor is white, this algorithm yields a correct estimate. For spatially correlated noise,  algorithms exploiting temporal standard deviation should be applied  \cite{buades2017denoising}. 
Since noise is independent for different images, such algorithms provide a more accurate estimate for correlated noise.  In our case, we prefer the first strategy since does not involve any registration stage, making the process faster. The algorithm is applied independently to each of the four channels.

The algorithm yields a set of observations $\{ x_i, \sigma_i\}$ which has to be interpolated to the whole range in order to have a complete noise model. 
Noise at the sensor is often assumed to follow a Poisson distribution, with linear variance.  However, this simplification is not valid for dark values where other sources of noise are dominant. In Figure \ref{fig:noiseint}, we illustrate how the obtained set $\{ x_i, \sigma_i\}$  is not well fitted by a single linear model. Instead, we preferred to use two different linear models, one for the dark and another one for lighter parts of the color range. We use a minimum least square method for optimizing such a model. We ask the two linear models to be jointly continuous, and the point in the range for which the approximation changes is also considered as a part of the optimization problem.  Figure \ref{fig:noiseint} compares the single linear approach to the proposed one. It is observed how this interpolation permits to correctly fit the noise model in the whole range.  A quadratic model could be used for fitting the darker part, but in such case we  should require the coefficient of degree 2 to be negative. We kept the simplest piecewise linear model.

\subsection{Variance stabilization}

Since noise at the CFA is signal dependent  we may apply a variance stabilization transform. We usually refer to the Anscombe transform as  the transformation $f(u) = 2 \sqrt{u + \frac{3}{8}}$ which is known to stabilize the variance of a Poisson noise model.  However, any signal dependent additive noise can be stabilized by a simple transform.
Let  $v  = u+ g(u) n$ be the noisy signal, we search for a function $f$ such that $f(v)$ has uniform standard deviation. When the noise is small compared to the signal we can apply the decomposition
$f(v) = f(u) + f'(u) g(u) n.$ Forcing  the noise term to be constant, $f'(u) g(u) = c$, and integrating we obtain
$$f(u) = \int_0^u \frac{c \, dt}{g(t)}.$$ When a linear variance noise model is taken, this transformation gives back  the known Anscombe transform.   The inverse transform is applied back after denoising to get the original range.  

\begin{figure*}[t]
\centering
\includegraphics[height=4 cm]{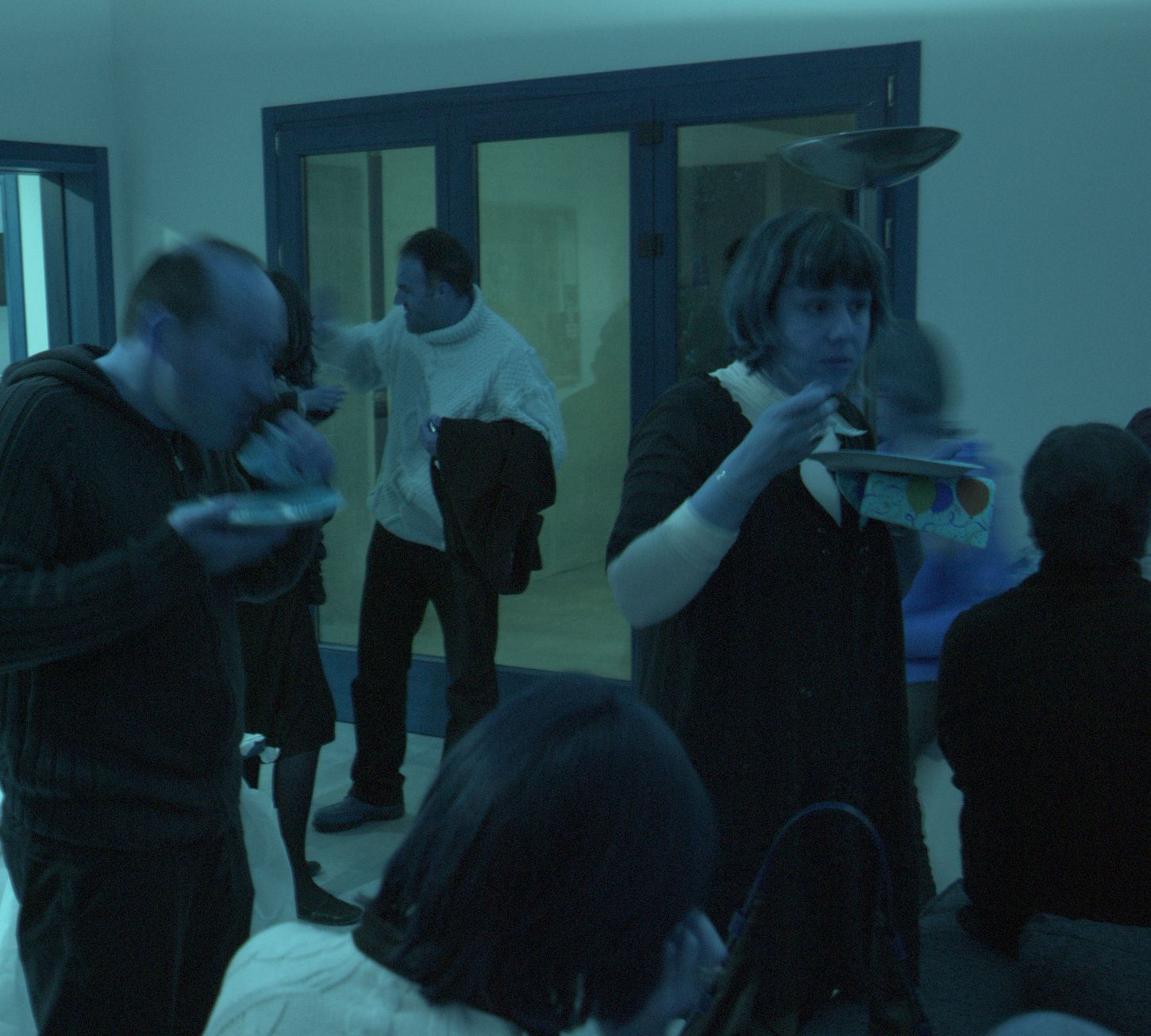}
\includegraphics[height=4 cm]{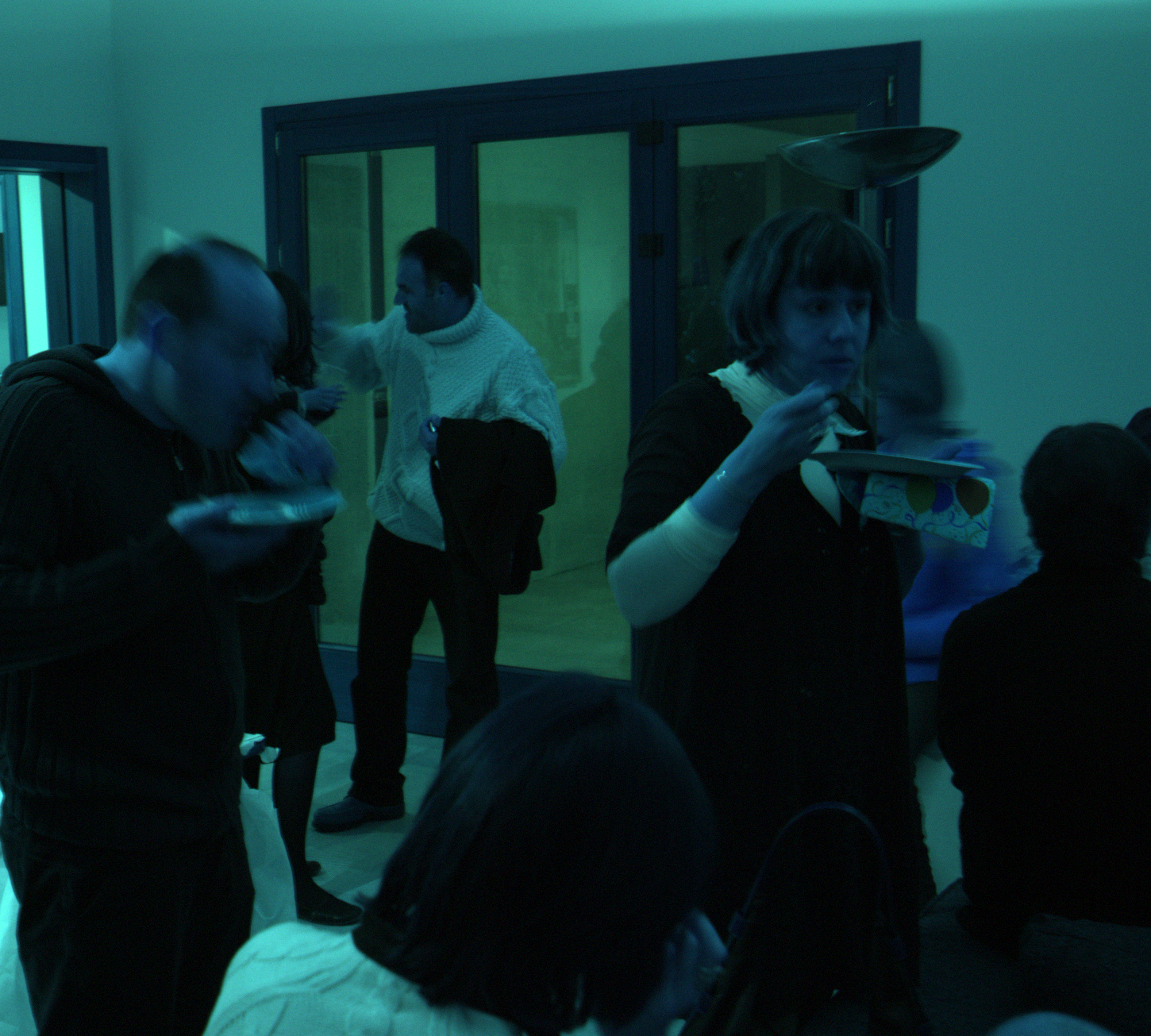}
\includegraphics[height=4cm]{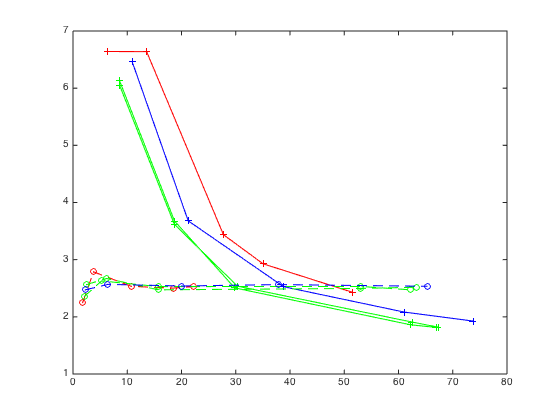}
\caption{Anscombe analysis.  We apply the classical anscombe and proposed variance stabilization transform to a CFA image.  Image is first sampled to be a full color image. Left: image after classical anscombe. Middle: image after proposed stabilization. Right: noise estimation in stabilized images. The dashed line (proposed estimation) is clearly more constant the continous line (classical anscombe) } \label{fig:anscombecomparison}
\end{figure*}

In  Figure \ref{fig:anscombecomparison} we compare the application of the classical Anscombe transform with the proposed variance stabilization. In order to do so, we estimate the noise for a single Nikon CFA image with the proposed approach. In order to apply the proposed variance stabilization, we set the parameter $c$, the amplitude of the noise after stabilization in such a way the color range of the transformed image  equals the range of the stabilized with the classical transform. In this way, we can now estimate the noise in these transformed images. We use the signal dependent noise estimation \cite{colom2014nonparametric} with 5 bins, and display this estimation in the same figure.  We appreciate, how noise amplitude is constant for the whole range after the proposed stabilization, which is not the case of the classical one. 

Since noise amplitude is different, we apply for each channel its own estimation, which is not the case of the classical anscombe which applies the same transformation for all channels.

\subsection{Noise removal} 

We will adapt the image sequence denoising method  in \cite{vdenoisingTIP15}.   We summarize the complete algorithm for denoising a frame $I_k$ from a sequence $\{I_1, I_2, \cdots, I_N \}$.  The same procedure is applied sequentially to all the frames of the sequence. 

First, the optical flow between $I_k$ and adjacent frames in a temporal neighborhood is computed and used for warping these frames onto $I_k$.  Occlusions are detected depending on the divergence of the computed flow: negative divergence values indicate occlusions.  Additionally, the color difference is checked after flow compensation. A large difference indicates occlusion, or at least failure of the color constancy assumption. 

Once the neighboring frames have been warped, the algorithm uses a 3D volumetric approach to search for similar patches, while still 2D image patches are used for denoising.  For each patch $P$ of the reference frame $I_k$, the  patch ${\cal P}$ referring to its extension to the temporal dimension is considered, having $M$ times more pixels than the original one (assuming $M$ patches in the temporal neighborhood). Since the images have been resampled according to the estimated flow,   the data is supposed to be static.  The algorithm looks for the $K$  extended patches closest  to ${\cal P}$.  As each extended patch contains $M$ 2D image patches, the group contains $K \cdot M$ selected patches. The Principal Component Analysis (PCA) of these patches is computed and their denoised counterparts are obtained by thresholding of the coefficients.   As proposed in \cite{zhang2010two}, the decision of canceling a coefficient is not taken depending on its magnitude, but the magnitude of the associated principal value.  A more robust thresholding is obtained by comparing the principal values to the noise standard deviation and canceling or maintaining the coefficients of all the patches associated to a certain principal direction.  The whole patch is restored in order to obtain the final estimate by aggregation.

Color denoising is achieved by considering color-patches with three times more samples than in the gray case. This permits to adapt both to geometry and color correlation among selected patches. However, this strategy has the disadvantage of being quite slow.   In our case, for a set of patches of $n \times n$ pixels,  this involves the computation of the PCA for vectors of size $4n^2$.  In addition in order to correctly compute such a representation we need the set of patches to be larger than the dimension of the vectors.  Both factors makes the algorithm slow for the removal of noise for a sequence of nowadays camera resolutions.  The method in \cite{vdenoisingTIP15}  also involves a second oracle step, in which the first denoised sequence is used to drive a second iteration of the method. We drop this second iteration in order to reduce the time of computation.

We will use a decorrelation transform YUVW for each 4-channel image.   In order to estimate this transform, we took several CFA images, converted to 4-channel ones and consider for each pixel the four dimensional vector containing its  red, two green and blue values. 
The PCA analysis of this set yields the transform
$$M=\left(\begin{array}{cccc} 0.5 & 0.5 & 0.5 & 0.5\\ -0.5 & 0.5& 0.5& -0.5 \\ 0.65 & 0.2784 & -0.2784&-0.65 \\-0.2784  & 0.65& -0.65& 0.2784 \\ \end{array}\right),$$
where each row of the matrix represents a principal vector. 
Similar analysis to standard color images yields an orthonormal version of the usual YUV color space.   
Since this is matrix an orthonormal matrix, its application after the variance stabilization transform makes the new components to have uniform noise with the same  standard deviation. The transposed matrix yields the inverse transform.

For each new channel YUVW, the sequence is denoised separately.  However, the component $Y$ is used for registering the frames and to guide patch selection  for all channels. After all components have been denoised, the inverse decorrelation transform is applied, as well as, the inverse variance stabilization transform to each channel.

\section{Demosaicking} \label{sec:demosaicking}

We introduce an image sequence algorithm to both interpolate missing values and remove residual noise from all pixels. 
The algorithm makes use of optical flow and exploits spatio-temporal redundancy with patch-based techniques to correct an initial interpolation. A decimation mask $D$ keeps the trace of initial CFA values and permits to differentiate from the initially interpolated ones. 

We apply this algorithm to generate a demosaicked and denoised video sequence from  $\{f_n=(\f^R_n,f^G_n, f^B_n)\}_{n=1}^N$  the mosaicked and eventually noisy frames.  The same strategy could be adapted for other filtering/interpolation tasks as video deinterlacing or the increase of temporal resolution.

\begin{figure*}[t]
\centering
\includegraphics[height=3.cm, width=3cm]{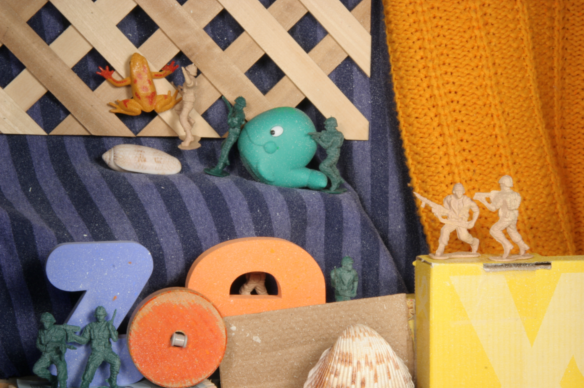}
\includegraphics[height=3.cm, width=3cm]{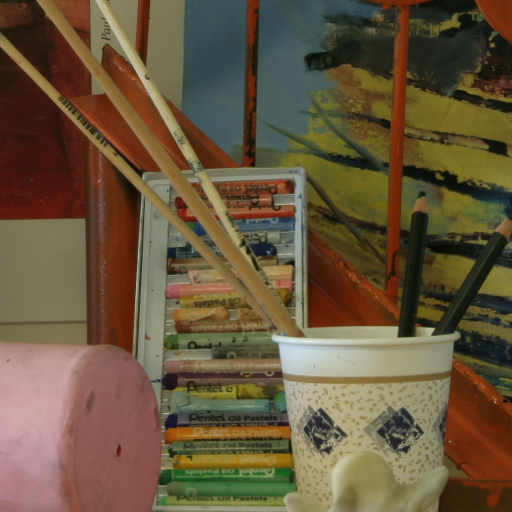}
\includegraphics[height=3.cm, width=3cm]{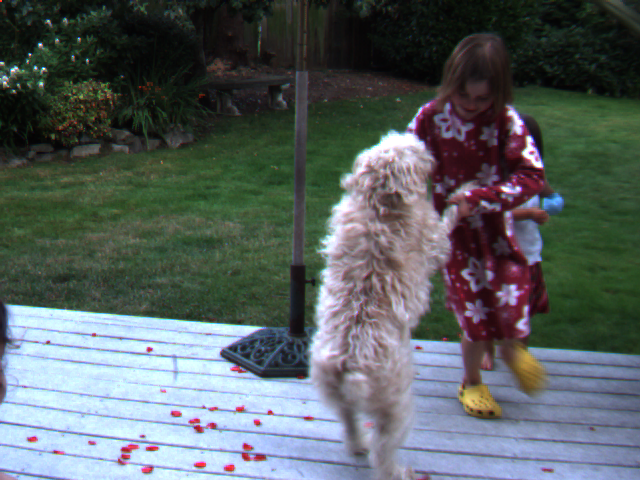}
\includegraphics[height=3.cm, width=3cm]{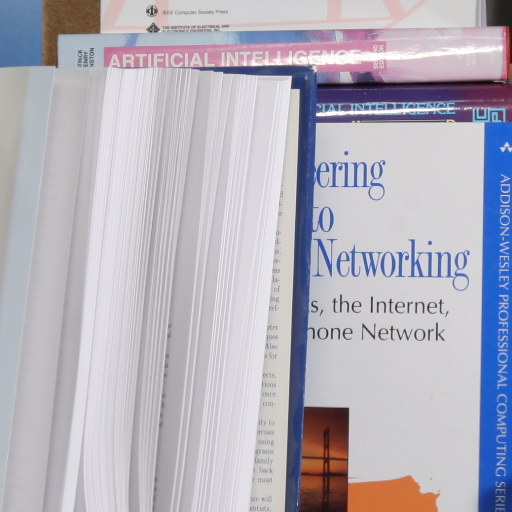}
\caption{Central view of each dataset used for simulated data for numerical comparison.} \label{fig:data}
\end{figure*}

\subsection{Spatio-temporal interpolation} \label{sec_interpolation}

Below we detail the algorithm to remove noise and aliasing from the initially interpolated sequence, denoted by $\{\tilde{f}_n\}_{n=1}^N$. 
This sequence might be obtained by bicubic interpolation, any anisotropic  or demosaicking method.
The algorithm describes the processing of a reference frame $\tilde{f}$.  Even if this description applies also to multivalued data, we will apply it to single channel images.  

In order to process $\tilde{f}$, we need to establish a correspondence function between  this frame and all the other ones in the sequence. We use the TV-L1 optical flow introduced in \cite{zach2007duality}. 

The goal is to remove the aliasing and increase the quality of the initially interpolated images $\tilde{f}_1, \ldots, \tilde{f}_N$ by weighted averaging. The algorithm proceeds patch per patch of the reference interpolated frame $\tilde{f}(P)$ by computing
\begin{equation} \label{eq:upsampling}
\hat{u}(P) = \frac{1}{C}  \cdot  \sum_{P_n^k \in \mathcal{N}_P}  w(\tilde{f}(P), \tilde{f_n}(P_n^k))   \   D(P_n^k) \cdot \tilde{f}_n(P_n^k),
\end{equation}
being $D$ the decimation mask associated to the sampling operator and $w(\tilde{f}(P), \tilde{f_n}(P_n^k))$ a real number measuring the similarity between patches $\tilde{f}(P)$ and $\tilde{f_n}(P_n^k)$. The number $n$ denotes the index of the image which  $P_n^k$ belongs to. In this setting, $C$ is a normalization factor and the operator $\cdot$ denotes the product element by element of each patch. The division by the normalization patch $C$ is also made element by element:
\begin{equation} \label{eq:normalization}
C =  \sum_{P_n^k \in \mathcal{N}_P}   w(\tilde{f}(P), \tilde{f_n}(P_n^k))\, D(P_n^k).
\end{equation}
The use of the decimation mask $D$, which is assumed to be the same for the sake of simplicity (the algorithm performs in a similar manner if the downsampling operator differs for each frame), makes the algorithm average only initial values.

The selection of candidate patches in $\mathcal{N}_P$ actually depends on a 3D distance taking into account motion estimation. This makes the selection procedure more robust to noise and aliasing artifacts. For each reference patch $\tilde{f}(P)$,  we denote as ${\mathcal P}$ its motion-compensated extension to the temporal dimension, having $N$ times more pixels than the original one:
\begin{equation*} 
\mathcal{P} = \bigcup_{n=1}^N  (P + W^{\top}_n(P)),
\end{equation*}
where $W^{\top}_n$ is a motion shift for patch $P$ and $n$-th frame. In practice, we take this shift to be the estimated forward flow between images $\tilde{f}$ and $\tilde{f}_{n}$ at the pixel in the centre of patch $P$.  

The algorithm looks for the $K$  extended patches closest  to ${\cal P}$ minimizing the distance
\begin{equation} \label{eq_interpol_w}
d(\mathcal{P}, \mathcal{Q} ) =  \sum_{n=1}^N\| \tilde{f}_n(P + W^{\top}_n(P) ) - \tilde{f}_n(Q + W^{\top}_n (Q))\|^2 .
\end{equation}
As each extended patch contains $N$ 2D patches, the selected group $\mathcal{N}_P$  contains $K \cdot N$ patches,  
\begin{equation*}
\left\{  P_n^k=P^k+W^{\top}_n(P^k)  \mid n=1, \ldots, N, \, k=1, \ldots, K \right\}.
\end{equation*}
We compute the similarity of each of these 2D patches and P  as
\begin{equation} \label{eq_interpol_w2}
w\left(\tilde{f}(P), \tilde{f}_n(P_n^k)\right) =  \exp\left(- \dfrac{ \| \tilde{f}(P)- \tilde{f}_n(P_n^k)  \|^2}{h^2}\right).
\end{equation}
The value of $h$ depends on the degree of aliasing and the noise statistics.  The preselection procedure using motion compensation makes this value less critical than in other patch-based regularization techniques \cite{ebrahimi2008multi,protter2009generalizing}. Finally, each pixel is estimated by aggregating all the values obtained by all patches containing it.

Importantly, no occlusion detection is performed on the estimated flow. In addition, as occlusion regions might be different from one frame to the other, it makes no sense to use the distance $d(\mathcal{P}, \mathcal{Q} )$ for computing the final weight. The comparison between patches $\tilde{f}(P)$ and $\tilde{f_n}(P_n^k)$ acts as a validation stage and avoids averaging very different patches.

\subsection{Initializations and flow estimation}

We first compute initial demosaicked images $\{\tilde{\f}_n\}_{n=1}^N$. For each frame, we use the local directional interpolation method proposed in \cite[Section II]{duran2014self}.  We apply this demosaicking algorithm to the output of our CFA denoising algorithm.  The initially demosaicked image may contain  typical demosaicking artifacts and eventually residual noise.

The method in \cite[Section II]{duran2014self} first estimates the green channel. At each red and blue position determined by the CFA, four interpolated values of the green are computed along north, south, east and west directions. Instead of deciding for each pixel which is the dominant direction and interpolate according to it, this decision is taken a posteriori once a full color image has been reconstructed for each direction. Therefore, for each of these directionally interpolated green channels, the red and blue components are reconstructed by bilinear interpolation on the channel differences green-red and green-blue. Finally, a decision of the most suitable approximation is made according to the variation of the chrominance (measured in the YUV space) at each pixel along the four directions.

Once the initial demosaicked video has been generated, we proceed to compute the optical flow between each pair of images in the sequence.  Due to the higher sampling rate of the green component which permits an easier reconstruction of the main geometry and texture than the red and  blue channels, the optical flow is computed on the sequence of interpolated green images $\{\tilde{f}^G_n\}_{n=1}^N$.

\subsection{Spatio-temporal demosaicking}

The green channel of each frame is first updated following the described spatio-temporal interpolation method applied to  $\{\tilde{f}^G_n\}_{n=1}^N$. The  mask $D$ used in \eqref{eq:upsampling}-\eqref{eq:normalization} is  the CFA mask corresponding to the green channel, i.e., a quinqunx of factor $2$ for each line and column. Therefore, only original green values are used in the weighted averaging. Furthermore, the patch-based Euclidean distances  are computed on the initially interpolated green sequence $\{\tilde{f}^G_n\}_{n=1}^N$.  

Once all green channels have been updated, we apply the same strategy independently to the red and blue channels. Following \cite{duran2014self}, we apply a non local average to  the channel differences red-green and blue-green instead of the red and blue themselves.  For the sake of simplicity, we shall describe only the process for the red channel.

We consider the red channels of the initially interpolated frames $\{\tilde{f}^R_n\}_{n=1}^N$ and we compute at each pixel the difference with the updated green, that is, $\{\tilde{f}^R_n-u^G_n\}_{n=1}^N$. We apply now the spatio-temporal interpolation method  by considering $\{\tilde{f}^R_n-u^G_n\}_{n=1}^N$ as the input grayscale sequence.  The CFA mask of the red channel is used as mask D. Furthermore, the patch-based Euclidean distances in \eqref{eq_interpol_w}-\eqref{eq_interpol_w2} are computed on the updated green sequence $\{u^G_n\}_{n=1}^N$ instead of the channel difference.
Once the method has been applied and the channel differences have been updated at each pixel by patch aggregation, the green value is added back to get the final red component. This process is performed on each frame, so we get the final red channels $\{u^R_n\}_{n=1}^N$.  

The application of the method to the green channels removes any residual noise kept at the CFA denoising,  as it does the posterior application to the channel differences of the red and blue  with the updated green.

\begin{figure*}[t]
\centering
\includegraphics[width=2.7cm]{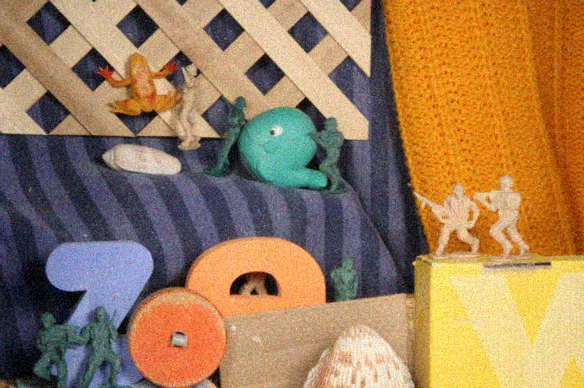}
\includegraphics[width=2.7cm]{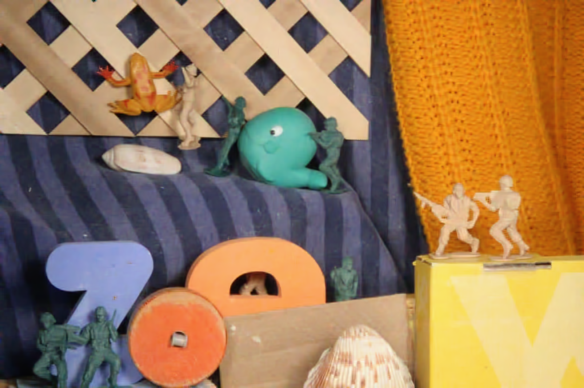}
\includegraphics[width=2.7cm]{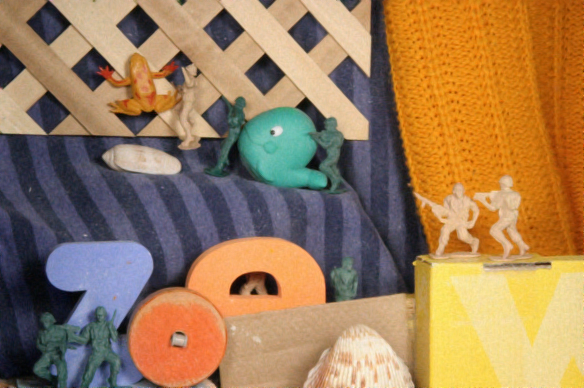}
\includegraphics[width=2.7cm]{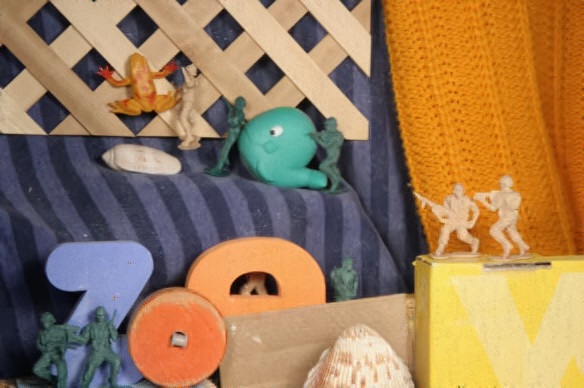}
\includegraphics[width=2.7cm]{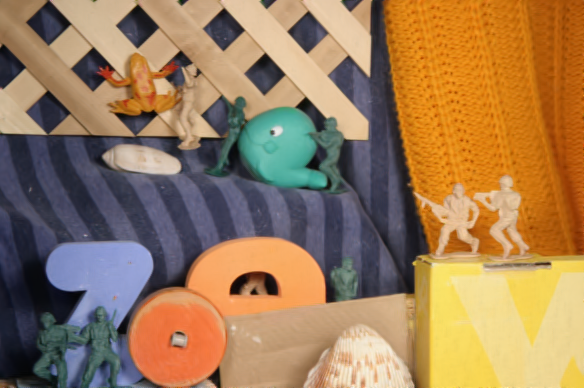}

\includegraphics[width=2.7cm]{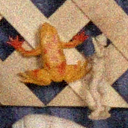}
\includegraphics[width=2.7cm]{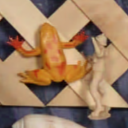}
\includegraphics[width=2.7cm]{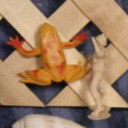}
\includegraphics[width=2.7cm]{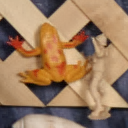}
\includegraphics[width=2.7cm]{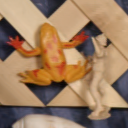}

\caption{Visual comparison with simulated noisy CFAs. From left to right:  image by local demosaicking, image by temporal demosaicking plus VBM3D, proposed demosaicking, proposed CFA denoising plus local demosaicking and complete chain. Below: detail of above images.} \label{fig:comparison}
\end{figure*}

\begin{figure*}[t]
\centering
\includegraphics[width=2.7cm]{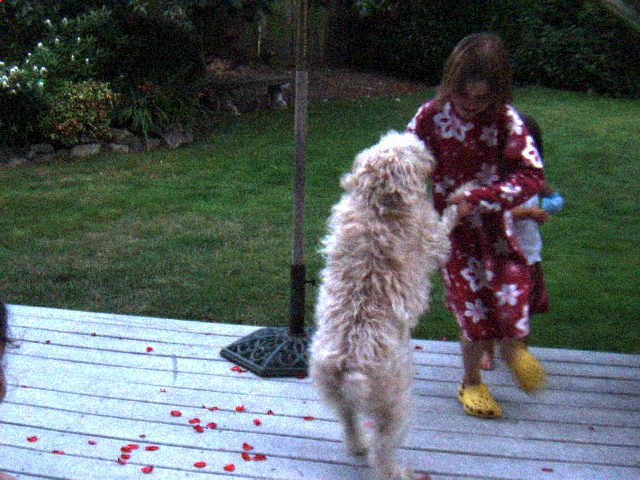}
\includegraphics[width=2.7cm]{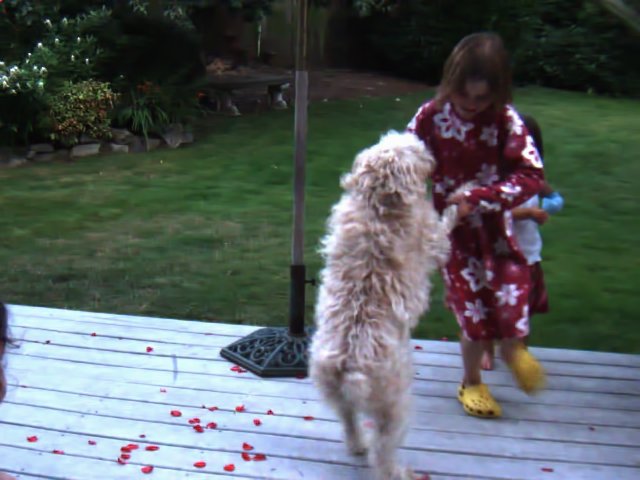}
\includegraphics[width=2.7cm]{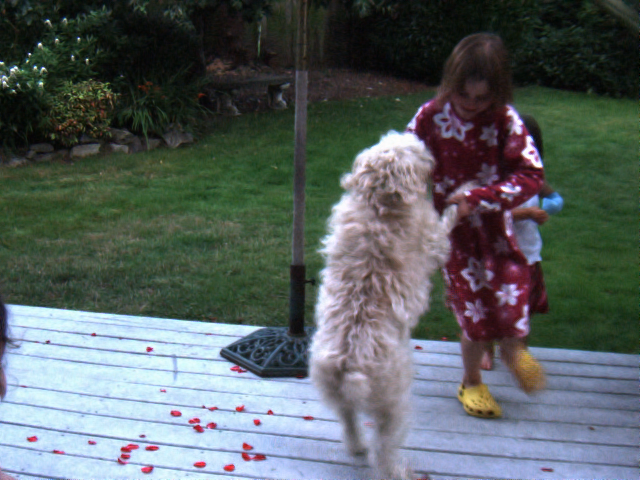}
\includegraphics[width=2.7cm]{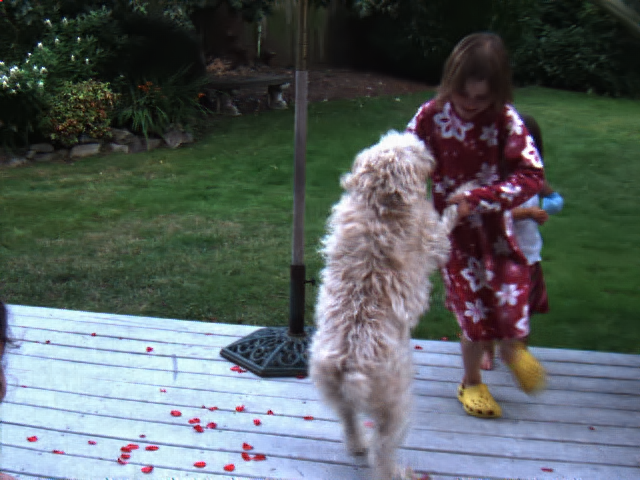}
\includegraphics[width=2.7cm]{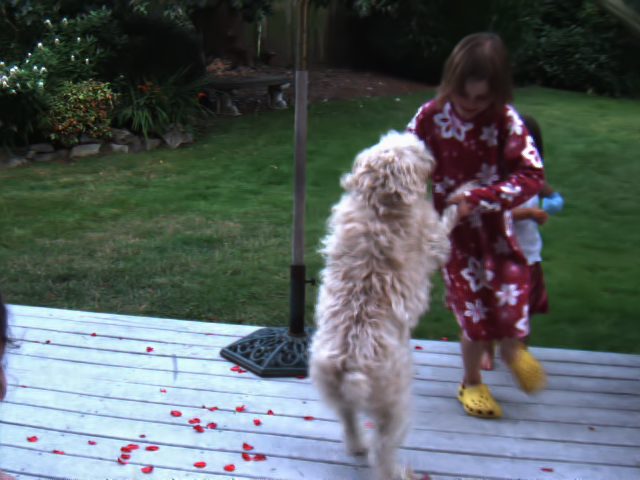}

\includegraphics[width=2.7cm]{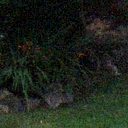}
\includegraphics[width=2.7cm]{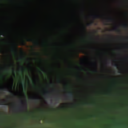}
\includegraphics[width=2.7cm]{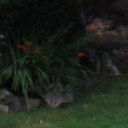}
\includegraphics[width=2.7cm]{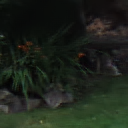}
\includegraphics[width=2.7cm]{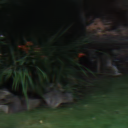}
\caption{Visual comparison with simulated noisy CFAs. From left to right:  image by local demosaicking, image by temporal demosaicking plus VBM3D, proposed demosaicking, proposed CFA denoising plus local demosaicking and complete chain. Below: detail of above images.} \label{fig:comparison2}
\end{figure*}

\section{Disussion and Experimental Results} \label{sec:experiments}

We test the performance of our method for both simulated and real data.   

\subsection{Simulated data}
We simulate noisy CFA sequences from a dataset of full color ones.  All sequences are composed of 8 frames, the central frame of each one is displayed in Figure \ref{fig:data}. We subsample the images according to the Bayer pattern and add noise of different standard deviations.   In this case, as images are already in color, we do not apply any particular imaging chain, thus colored spots are not enhanced by the white balance or gamma correction. Since we have the ground truth we evaluate the results both visually and numerically. 

We compare the proposed chain with,
\begin{itemize}
\item The local interpolation method introduced in \cite[Section II]{duran2014self}, which is used as initialization for the demosaicking stage without any denoising step.
\item The combination of the temporal demosaicking method introduced in \cite{wu2006temporal} with a posterior denoising using VBM3D \cite{dabov2007video}.  Since the standard deviation of noise is modified after the demosaicking process, we tested several parameters for VBM3D and kept the one with smallest error.   The VBM3D is a video restoration method grouping patches of consecutive frames and performing transform thresholding. It also has a second oracle iteration using a first estimate to drive the patch selection and thresholding.  Our algorithm does not involve any second iteration in order to reduce computation time.

\item The application of the proposed demosaicking strategy without any previous denoising. Since the demosaicking method modifies all values, CFA values inclusive, it actually removes noise. In such a case, the parameter $h$ is fixed depending on the noise standard deviation. 
\item The proposed CFA denoising with the posterior local interpolation  \cite[Section II]{duran2014self}.
\item The full proposed chain. In this case, the demosaicking algorithm only filters residual noise not completely removed by the denoising stage.
\end{itemize}

Table \ref{tab:results} displays the root mean square error (RMSE) for the central view of the sequence and noise standard deviations $\sigma=5$ and $10$.  The error of the local initialization is significantly larger than the other methods since it does not contain any denoising stage.  

For low values of $\sigma$, the proposed demosaicking method without denoising, improves the temporal demosaicking plus BM3D and the application of the proposed CFA denoising with the local demosaicking method. This illustrates the performance of the demosaicking method, and its robustness to noise.  For $\sigma=10$ the proposed demosaicking still performs better than the temporal demosaicking plus VBM3D. However,  the proposed CFA denoising with the single image local demosaicking outperforms the proposed demosaicking, as well as, the temporal demosaicking plus VBM3D. For both noise standard deviations, the proposed chain combining denoising and demosaicking largely outperforms the rest of algorithms.

\begin{table*}[t] \small
\centering
\begin{tabular}{|c|c|c|c|c|c|c|c|c|}
\hline
Image & Single Image  & Temporal dem.  & Prop. & Prop. Den  & Prop. \\ 
$\sigma=5$ & dem.  & plus VBM3D &  dem. & local dem. & chain  \\ \hline \hline 
{\it Army} & 5.39  & 3.55  & 3.07  & 3.80 & 3.16 \\\hline
{\it Art} & 4.98 & 3.36 & 3.23 & 3.2 & 2.91\\\hline
{\it Dog} & 5.14 &  8.08 &  8.19& 8.40 & 8.13\\\hline
{\it Books} & 9.31 &  3.48 &  3.48& 3.20 & 2.92 \\\hline \hline
average & 6.20 &  4.61 & 4.49   & 4.65 & 4.28\\\hline
\end{tabular}

\begin{tabular}{|c|c|c|c|c|c|c|c|}
\hline
Image & Single Image  & Temporal dem.  & Prop. & Prop. Den  & Prop.\\ 
$\sigma=10$ & dem.  & plus VBM3D & dem. & local dem & chain\\ \hline \hline 
{\it Army} &  9.50& 4.62 & 4.42 & 4.64 & 4.21 \\\hline
{\it Art} & 9.30 &  5.01 & 4.85& 4.32  & 4.01  \\\hline
{\it Dog} & 9.54 & 8.69 & 8.71  & 8.75 & 8.47  \\\hline
{\it Books} & 12.21 &  5.06  & 5.18& 4.21 & 3.87\\\hline \hline
average &  10.14 &  5.85 &  5.79 & 5.48 & 5.13\\\hline
\end{tabular}
%
%

\caption{Table \ref{tab:results} displays the root mean square error (RMSE) for the central view of the sequence and noise standard deviations $\sigma=5$ and $10$. } 
\label{tab:results}
\end{table*}

Figures \ref{fig:comparison} and \ref{fig:comparison2} display the results of each method for two sequences. The local demosaicking does not remove any noise. The images processed by the temporal demosaicking and VBM3D are slightly blurry and many details have been removed. Since noise is modified by the demosaicking schemes, its removal is quite challenging. The proposed demosaicking alone is able to correctly remove typical demosaicking artifacts, but is slightly noisy. The proposed chain is able to correctly demosaick and remove noise.

\subsection{Real data}

We test the proposed method with  real data.  
We use several image sequences acquired with a reflex camera and several RAW video sequences introduced in \cite{andriani2013beyond}. Since we need to have access to the RAW data, we could not acquire video ourselves but only images acquired consecutively with a Nikon D80. Some of these data are actually burst sequences, meaning images are acquired quasi instantaneously holding the camera in the same position. Since the camera is hand held, the view point and orientation of the camera slightly changes.  For these examples a standard imaging chain composed of white balance by gray mean, and a gamma correction with $\gamma=0.5$ are used. We do not apply any color correction, in order to convert the RGB values of the sensor to the one of the display. 

\begin{figure*}[t] 
\centering
\includegraphics[width=3.2cm]{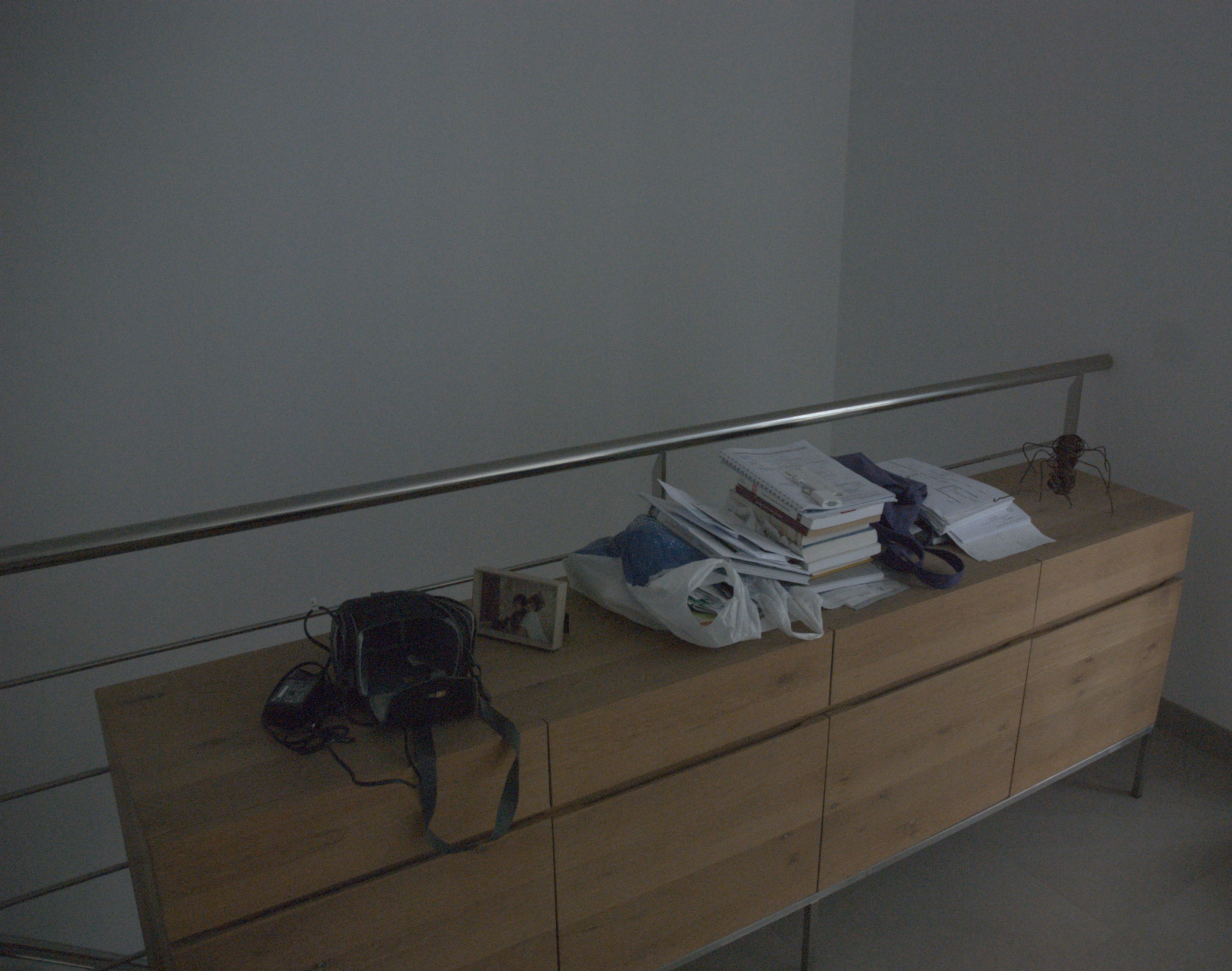}
\includegraphics[width=3.2cm]{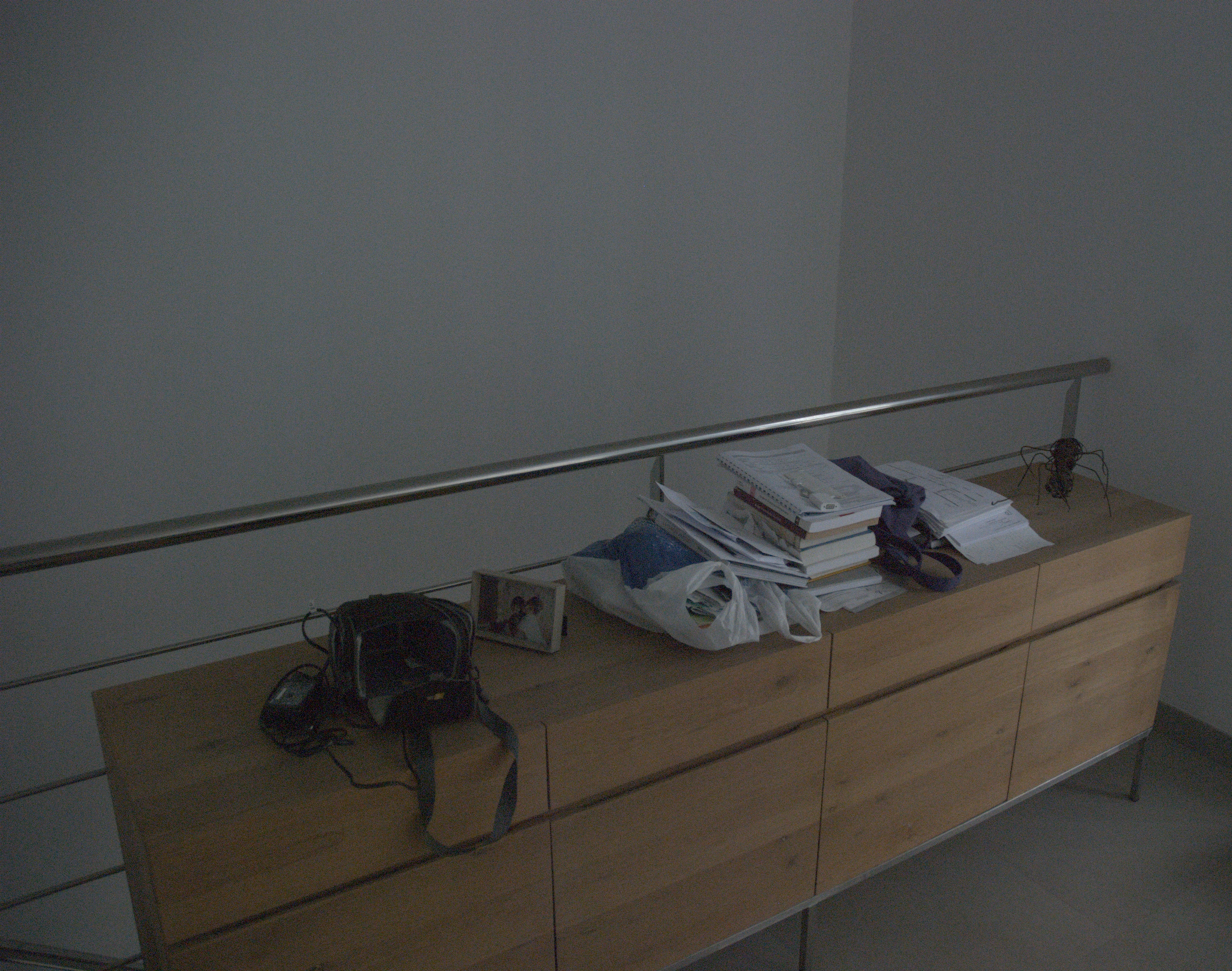}
\includegraphics[width=3.2cm]{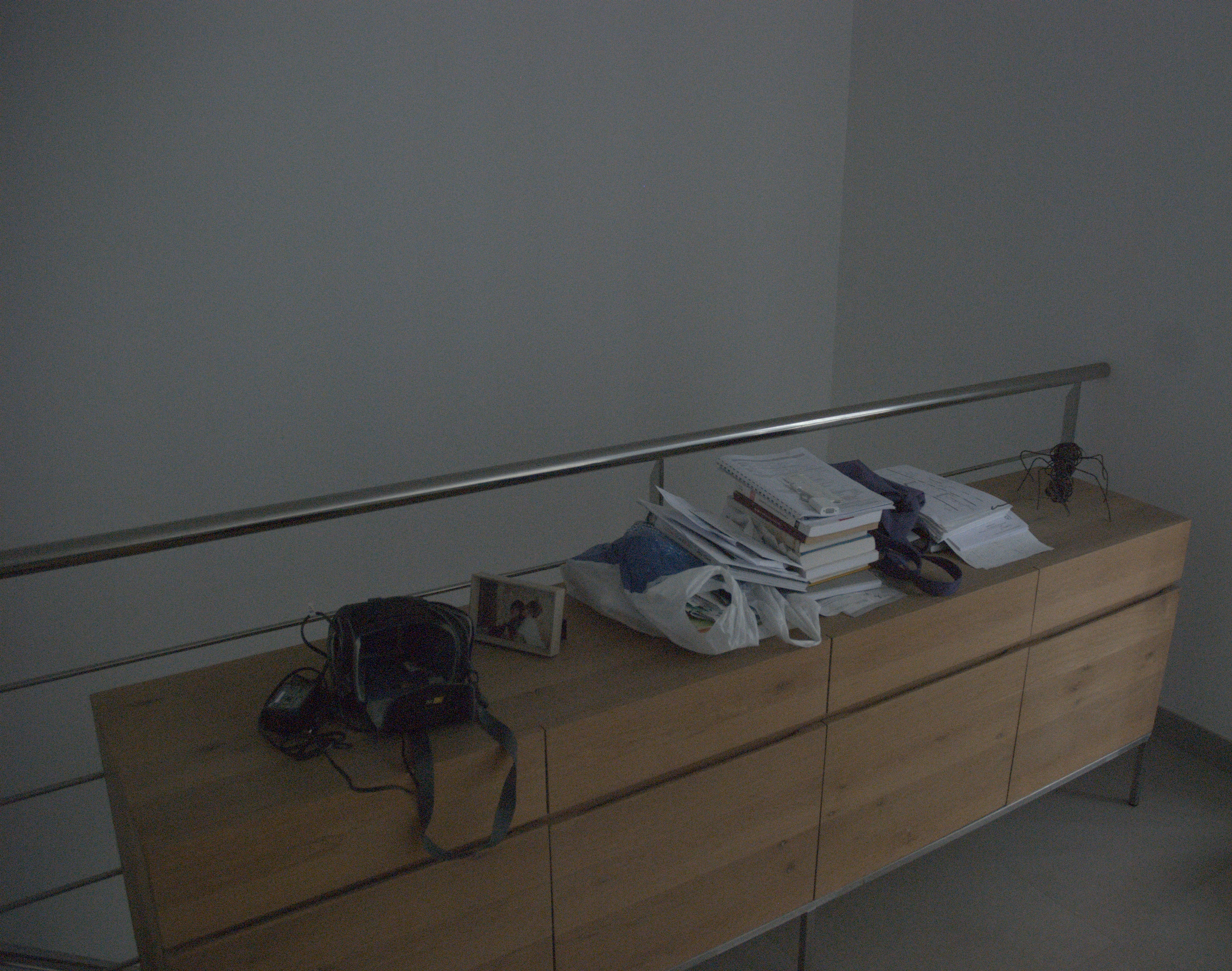}
\includegraphics[width=3.2cm]{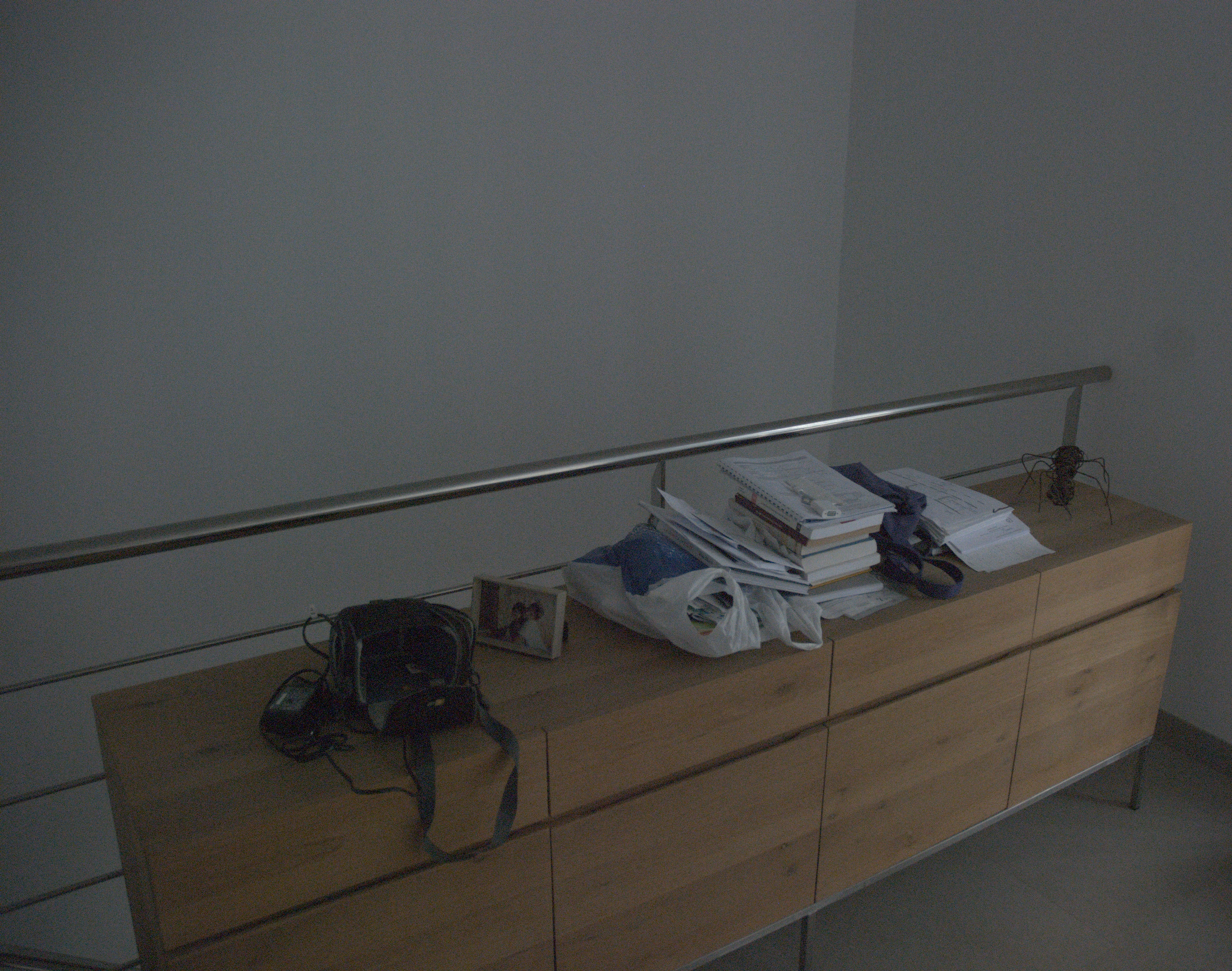}
\includegraphics[width=3.2cm]{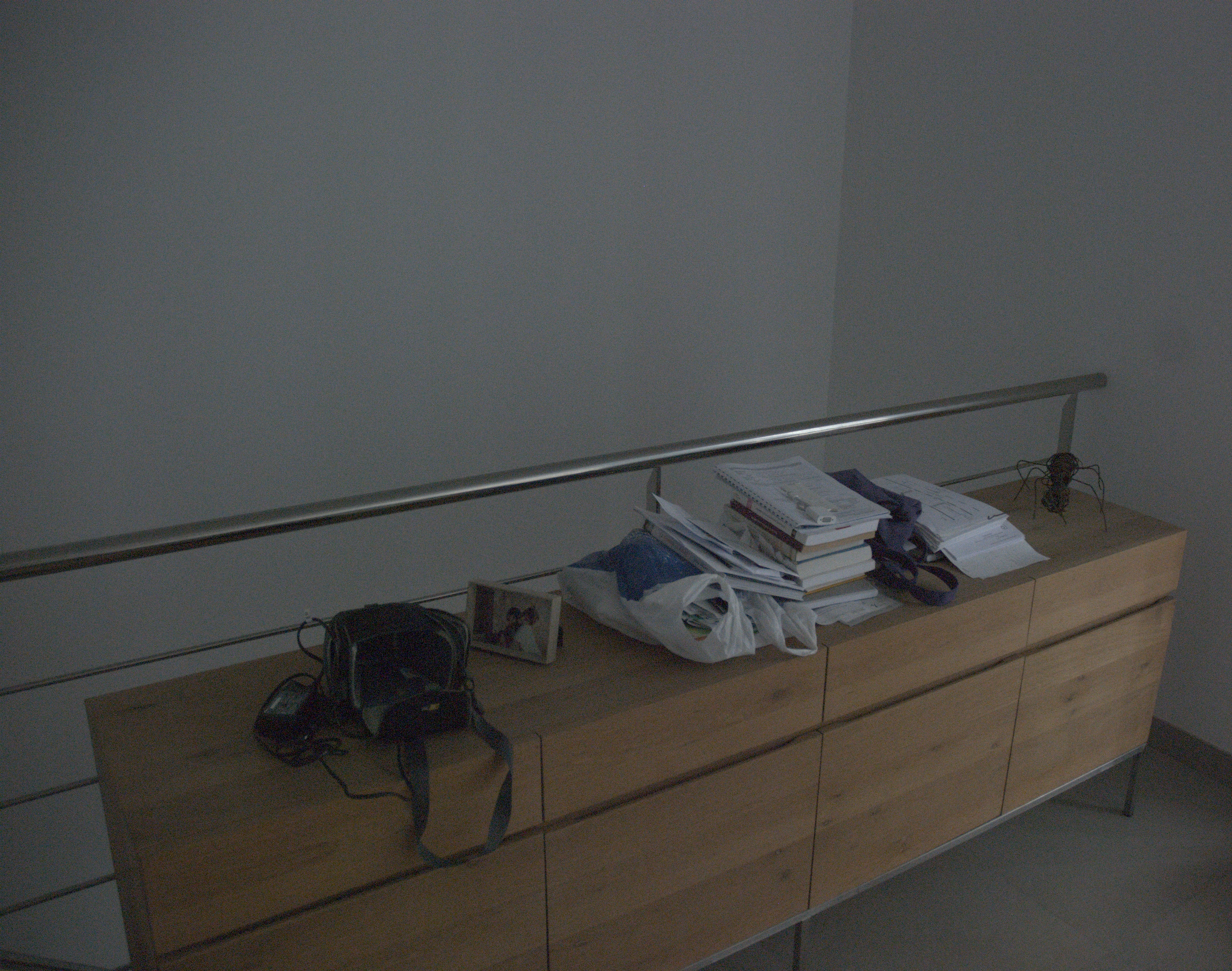}
\includegraphics[width=3.2cm]{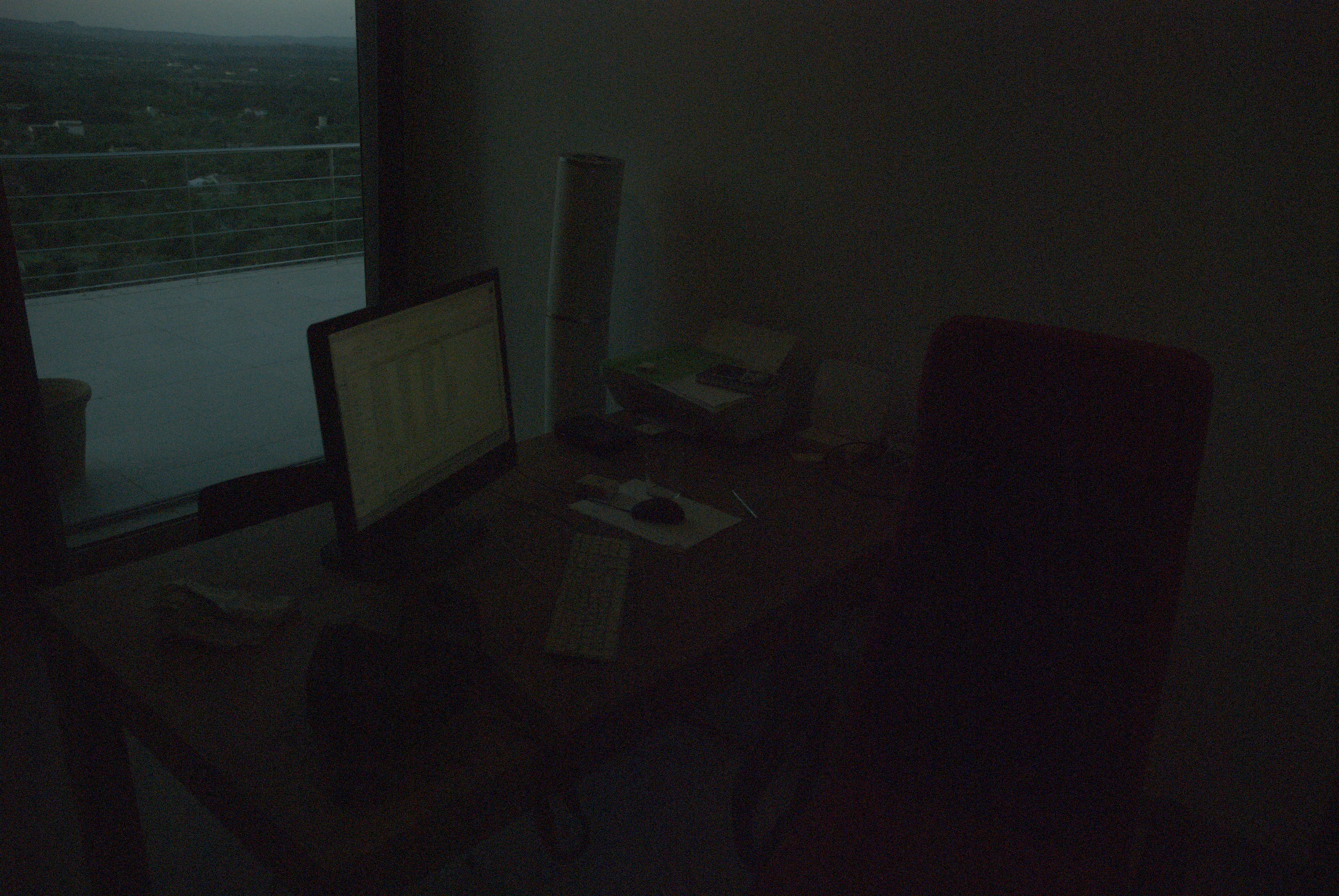}
\includegraphics[width=3.2cm]{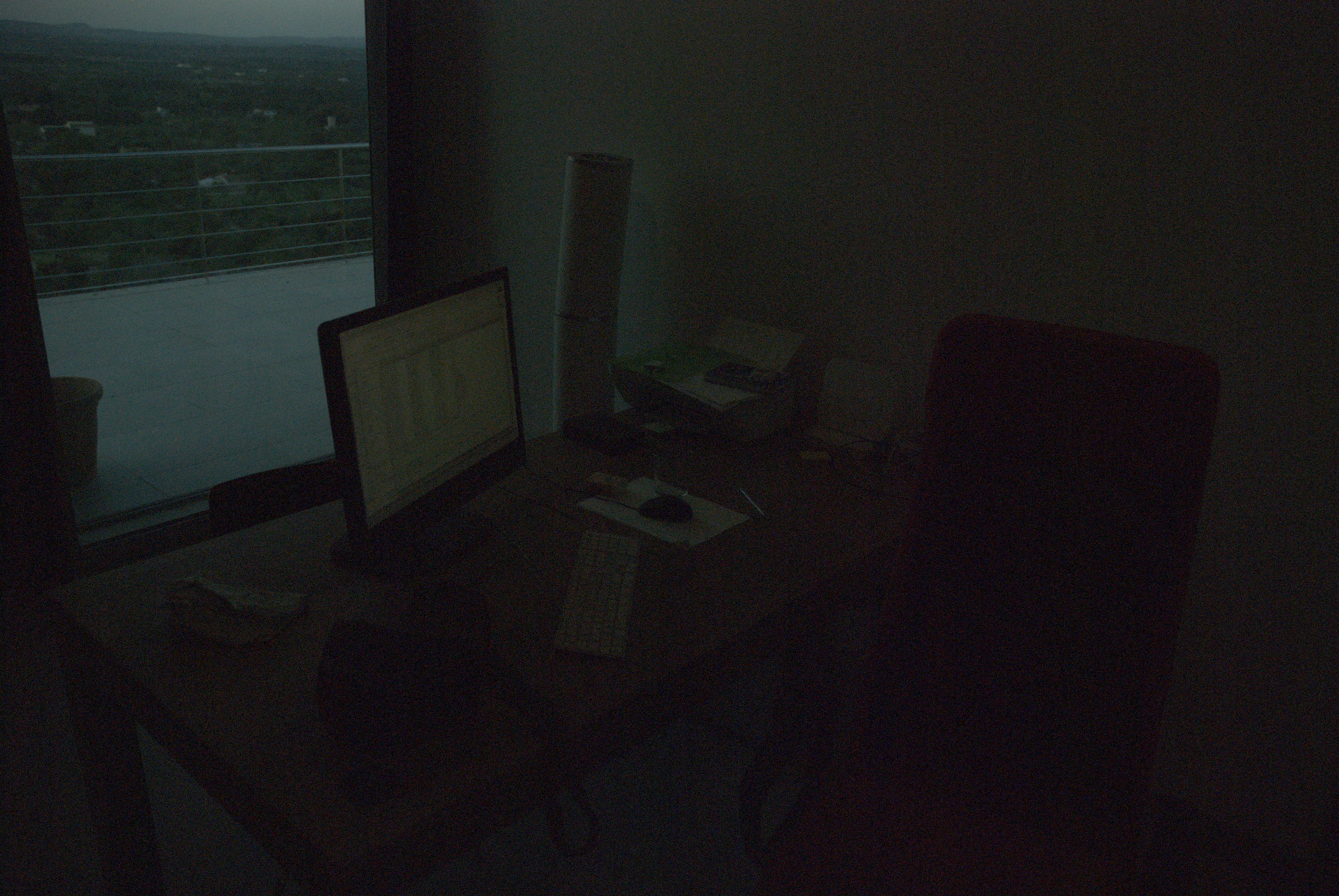}
\includegraphics[width=3.2cm]{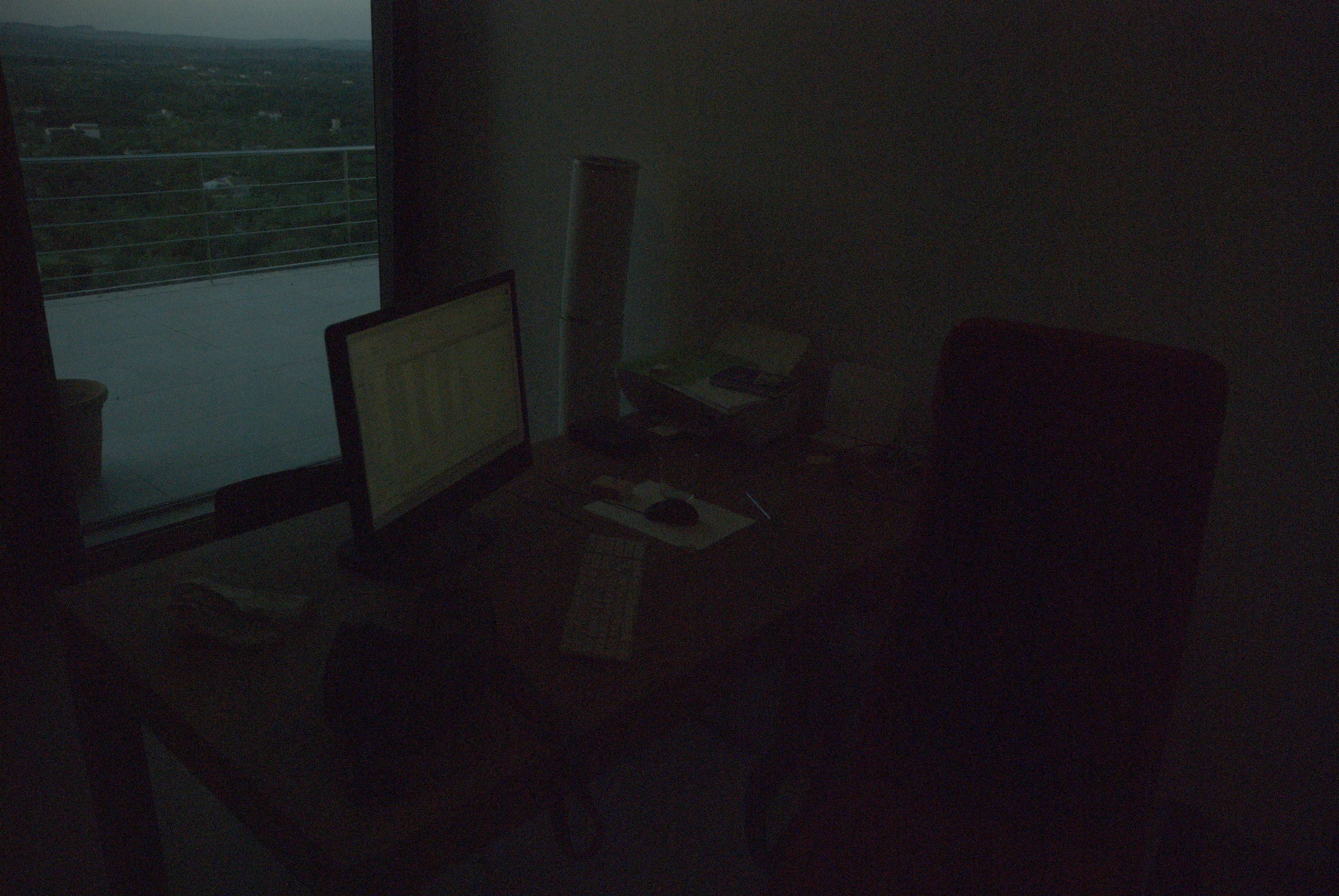}
\includegraphics[width=3.2cm]{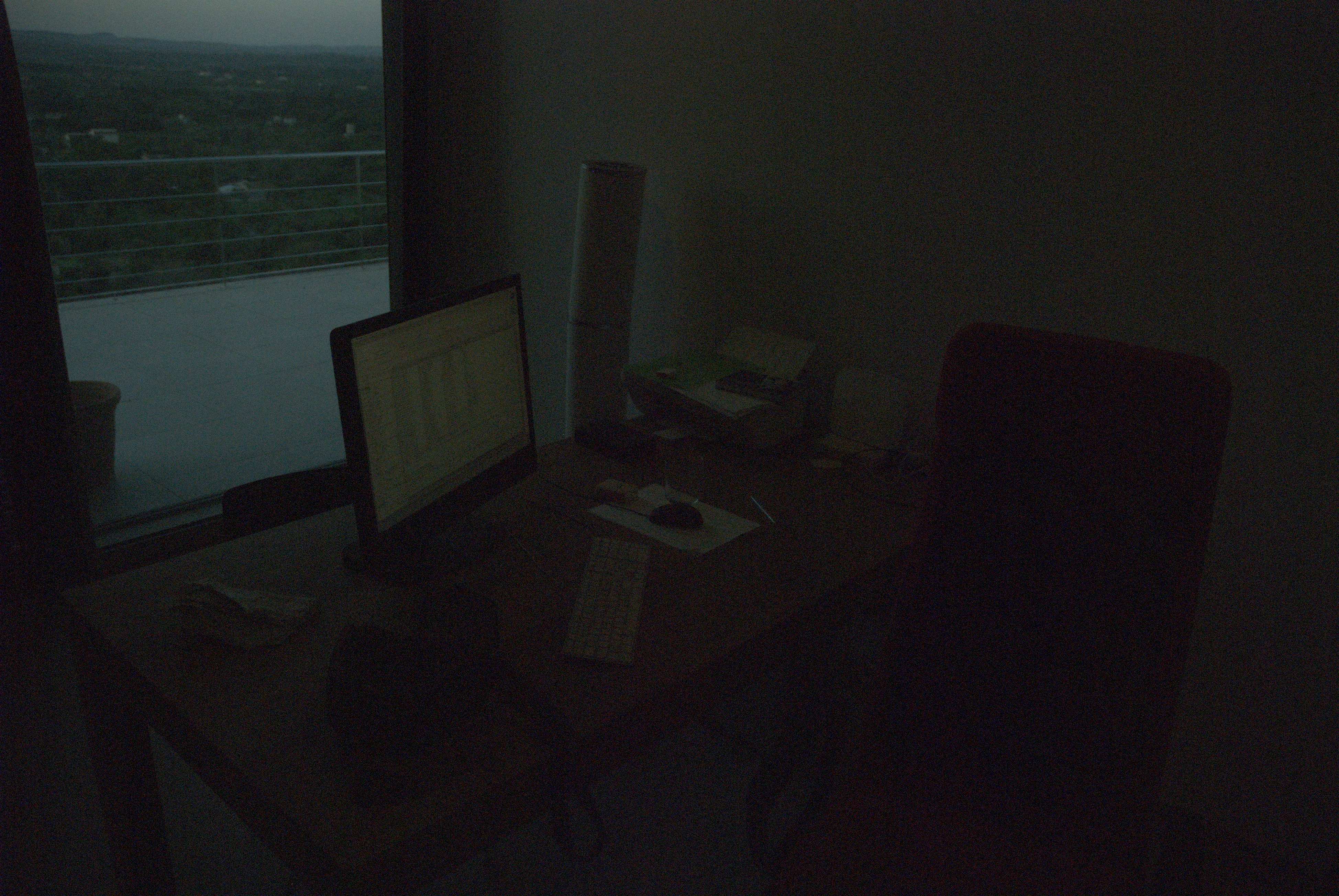}
\includegraphics[width=3.2cm]{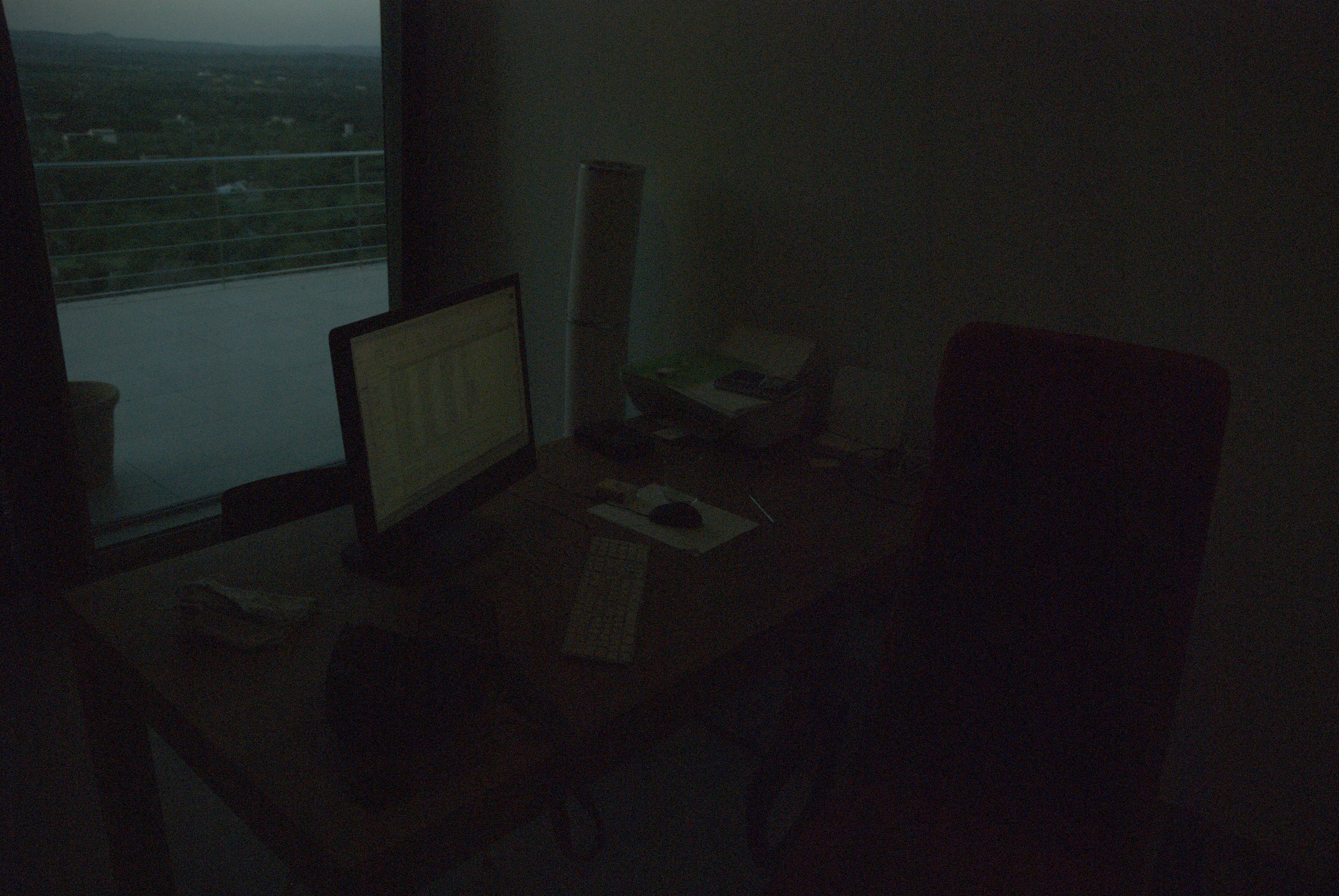}
\includegraphics[width=3.2cm]{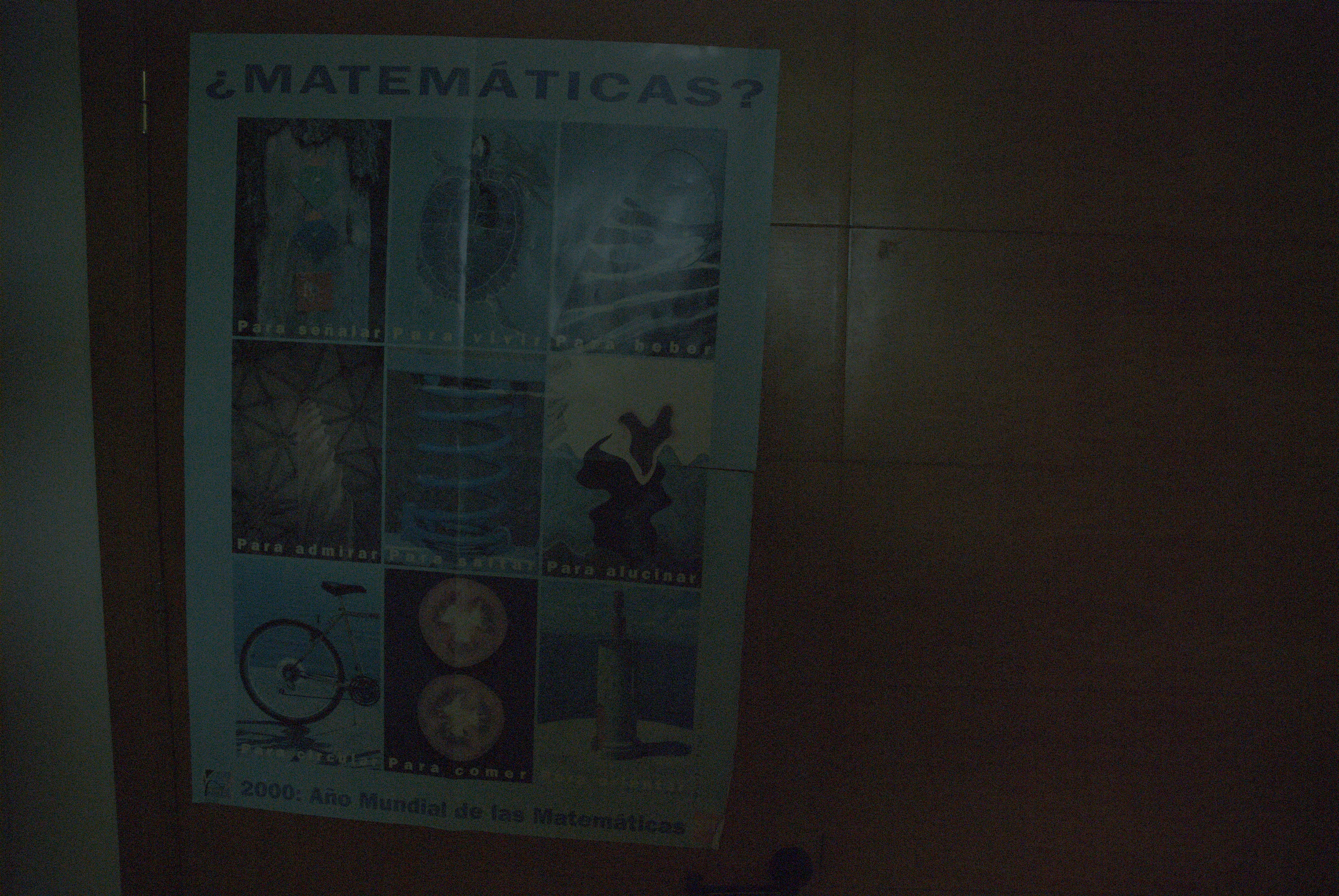}
\includegraphics[width=3.2cm]{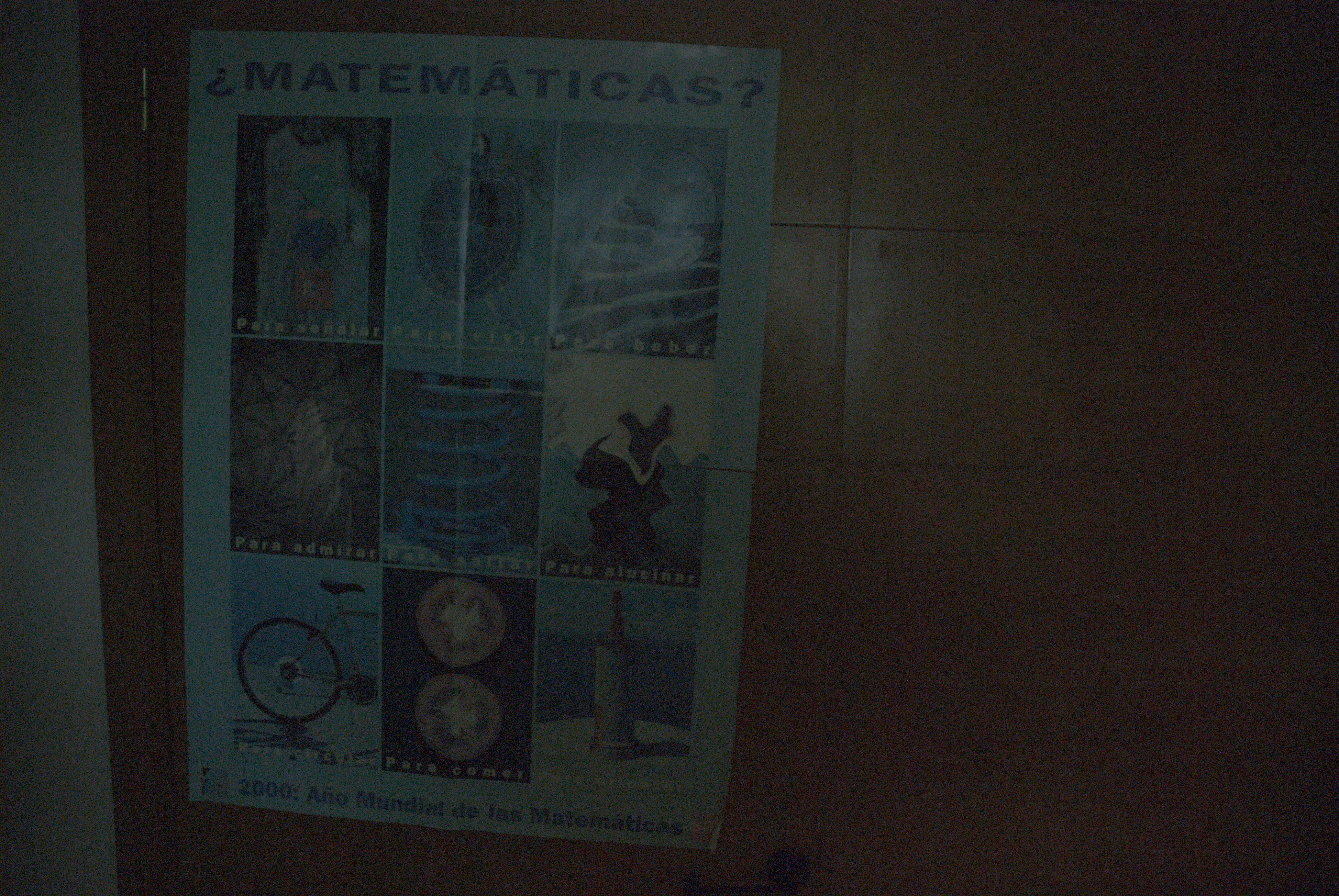}
\includegraphics[width=3.2cm]{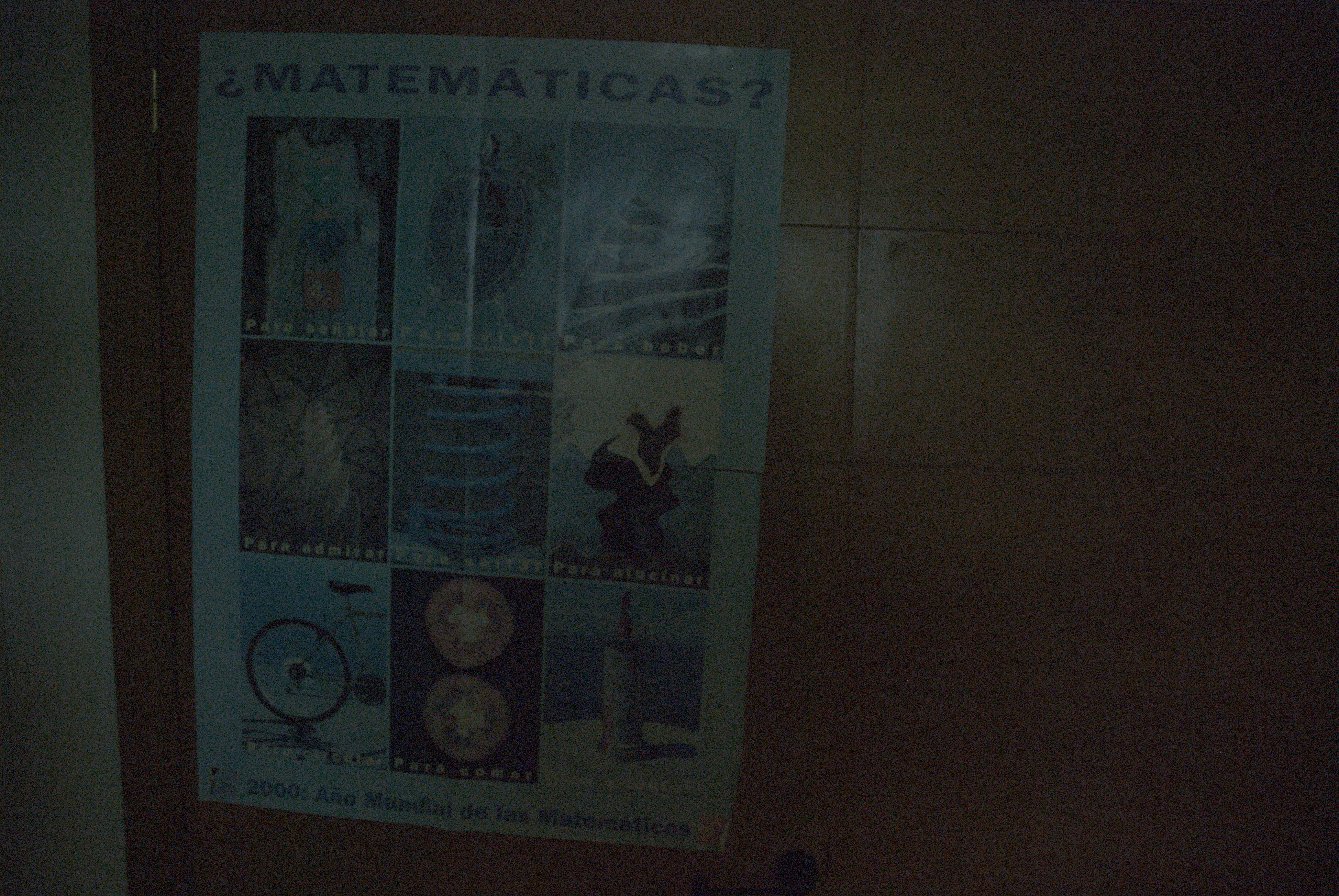}
\includegraphics[width=3.2cm]{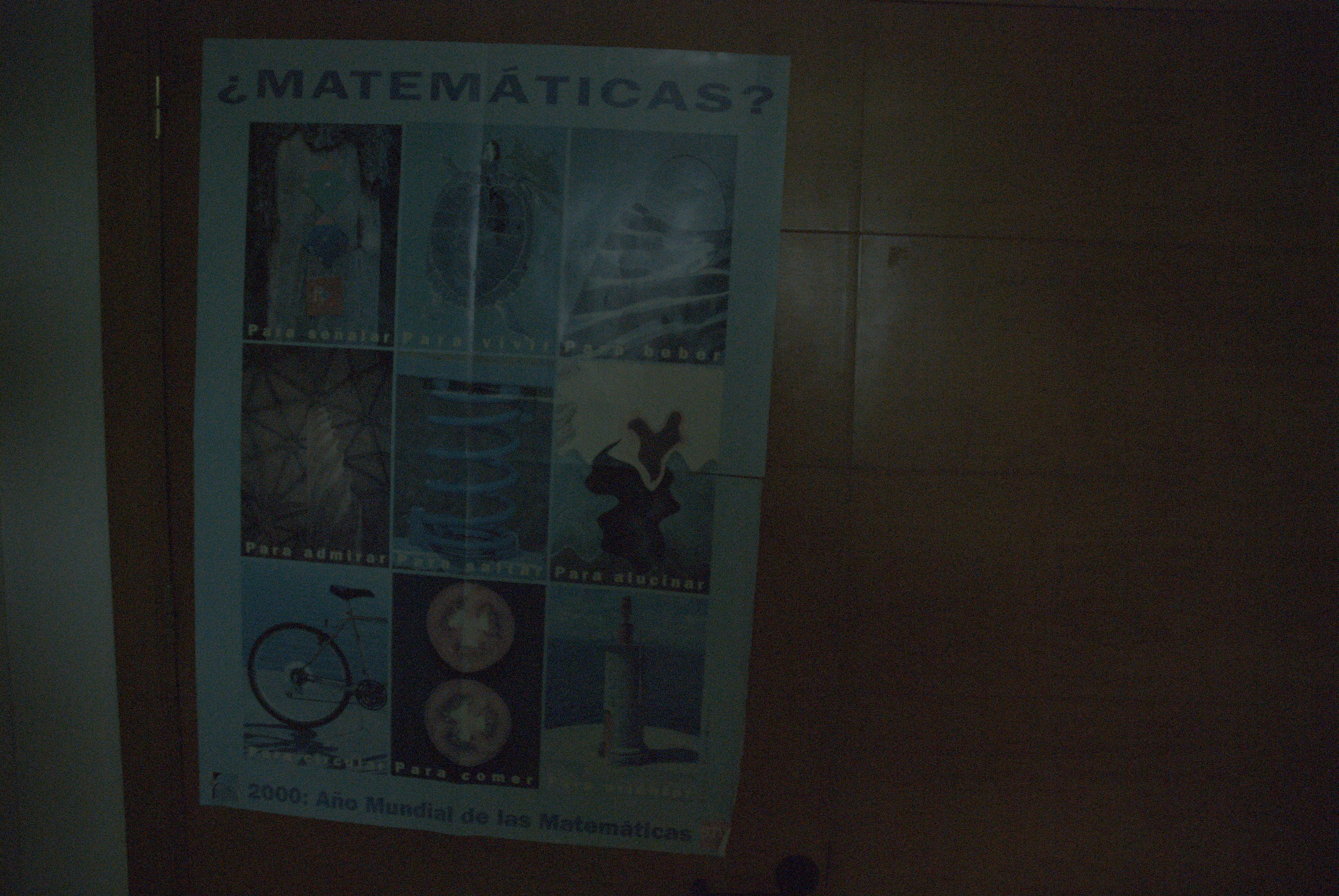}
\includegraphics[width=3.2cm]{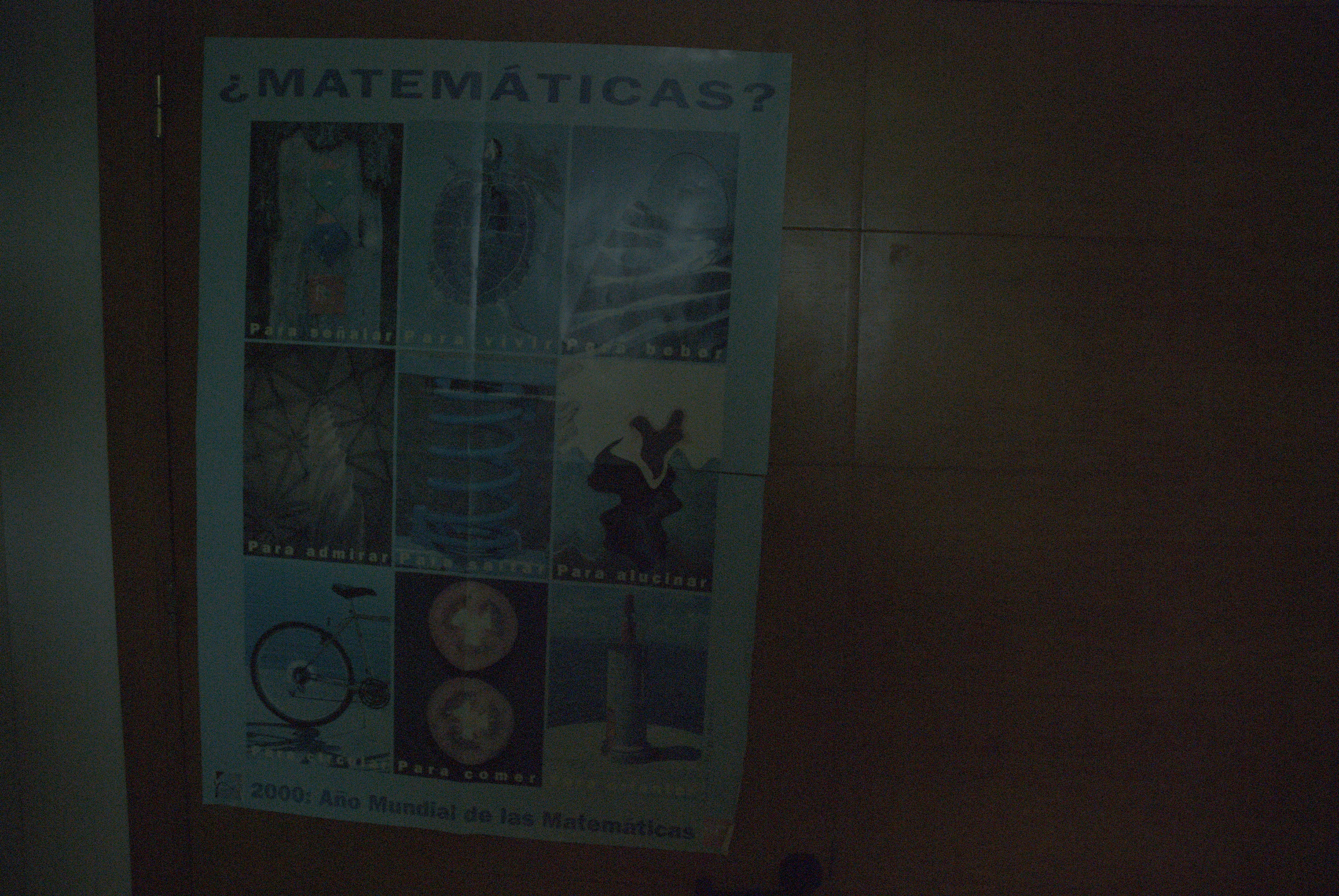}

\caption{We display several images of each sequence used in this experimentation section. First three rows were acquired in RAW format using a Nikon D80 with the same ISO, aperture and exposure time. We display the images after the demosaicking and corresponding imaging chain is applied. We remind that the RAW images will be used for processing.} \label{fig:dataRAW}
\end{figure*}

For real video, we use the video sequences in RAW format introduced in \cite{andriani2013beyond}.   The imaging pipeline applied after the demosaicking, for these videos, is also made available in the same publication. 
The sequences used in this section are displayed in Figure  \ref{fig:dataRAW}.  Demosaicking and the color chain have been applied for visualization purposes, although the RAW data is the one actually being used.

We compare the application of the standard imaging chain with local demosaicking, the proposed complete chain and the method  in \cite{buades2017denoising}. The method in   \cite{buades2017denoising}  removes noise after the imaging pipeline is applied. The method  performs in a multi-scale framework, where noise is estimated at each scale, and the video denoising method in \cite{vdenoisingTIP15} is used after variance stabilization. 

\begin{figure*}[t]
\centering

\includegraphics[width=5.2cm]{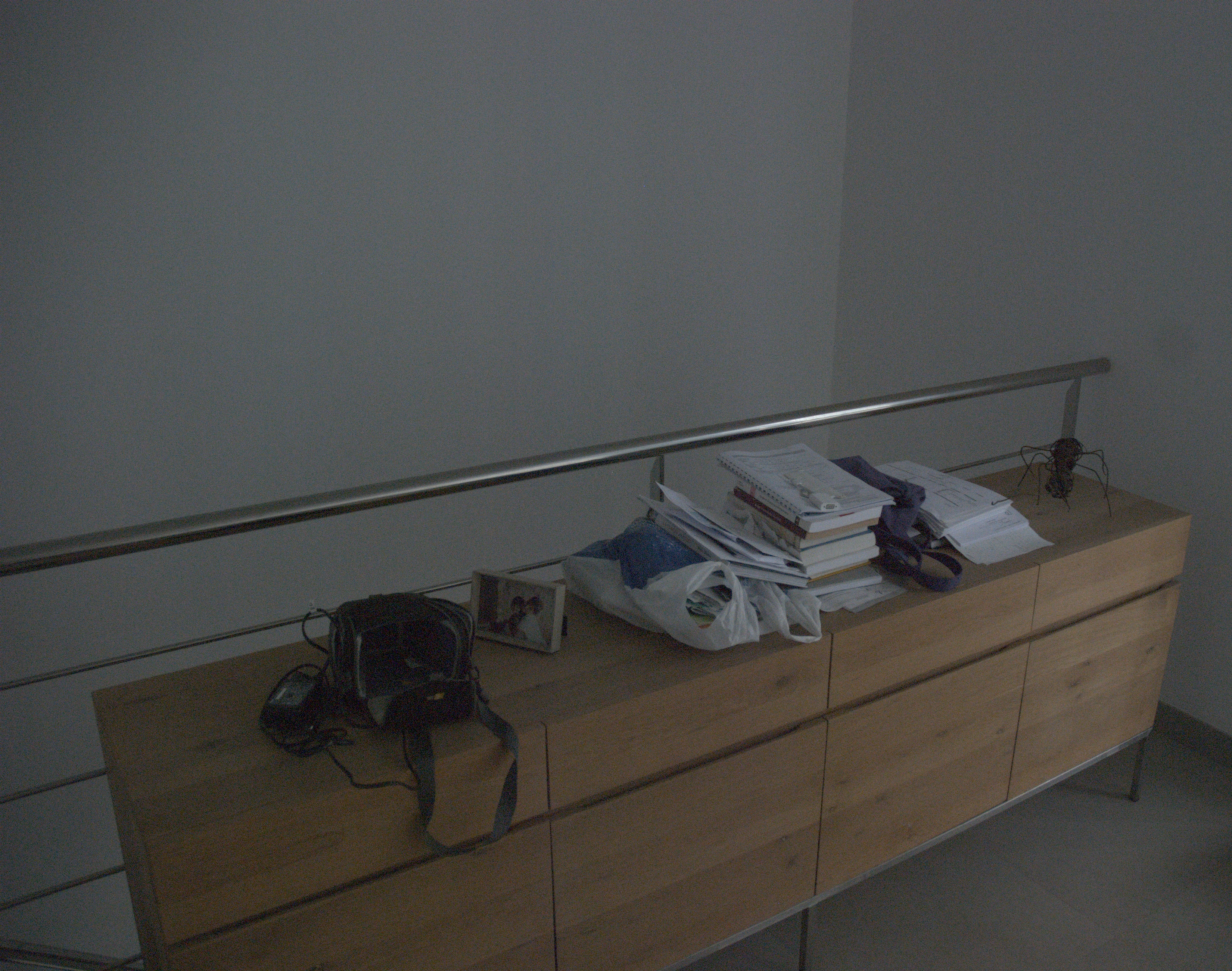}
\includegraphics[width=5.2cm]{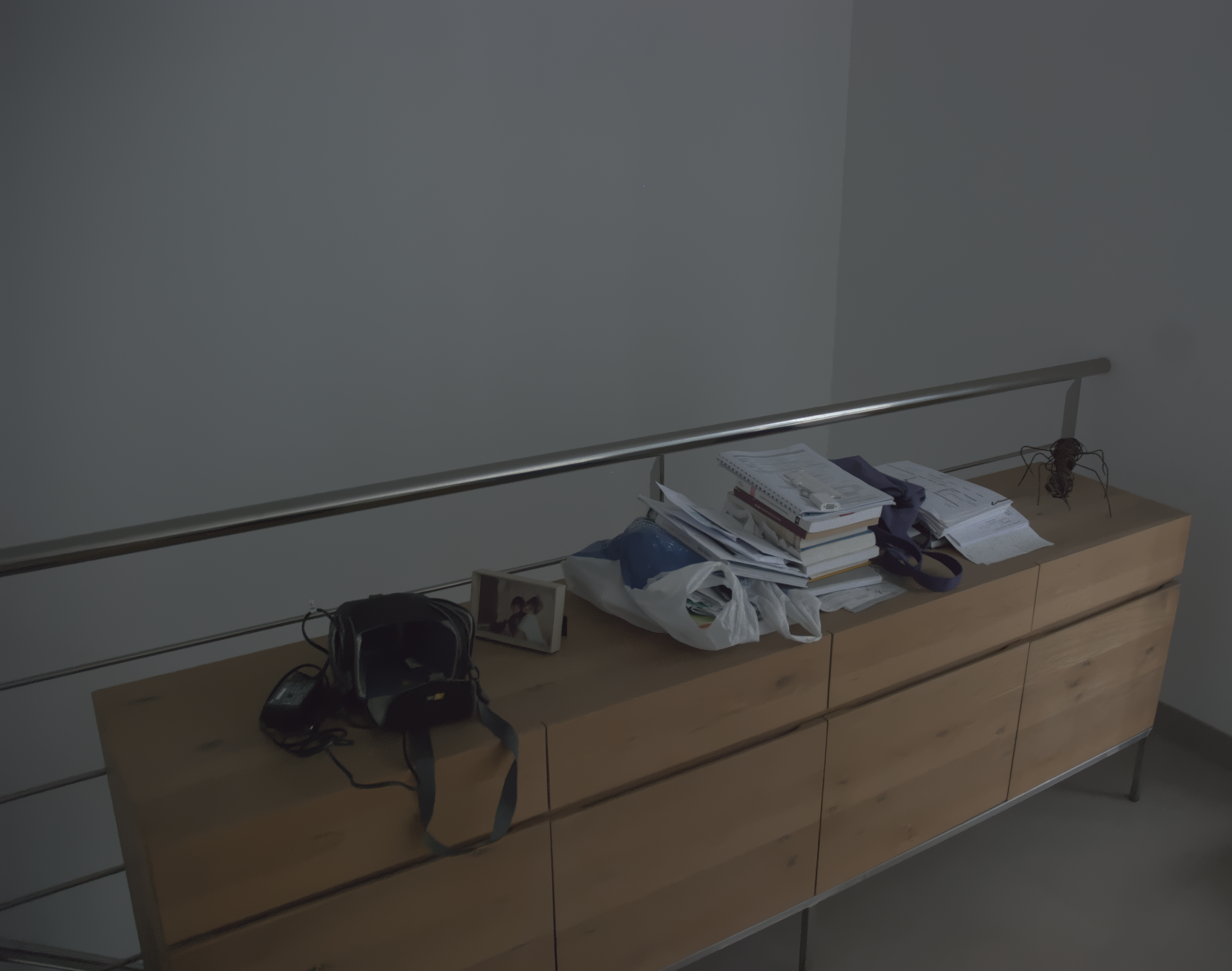}
\includegraphics[width=5.2cm]{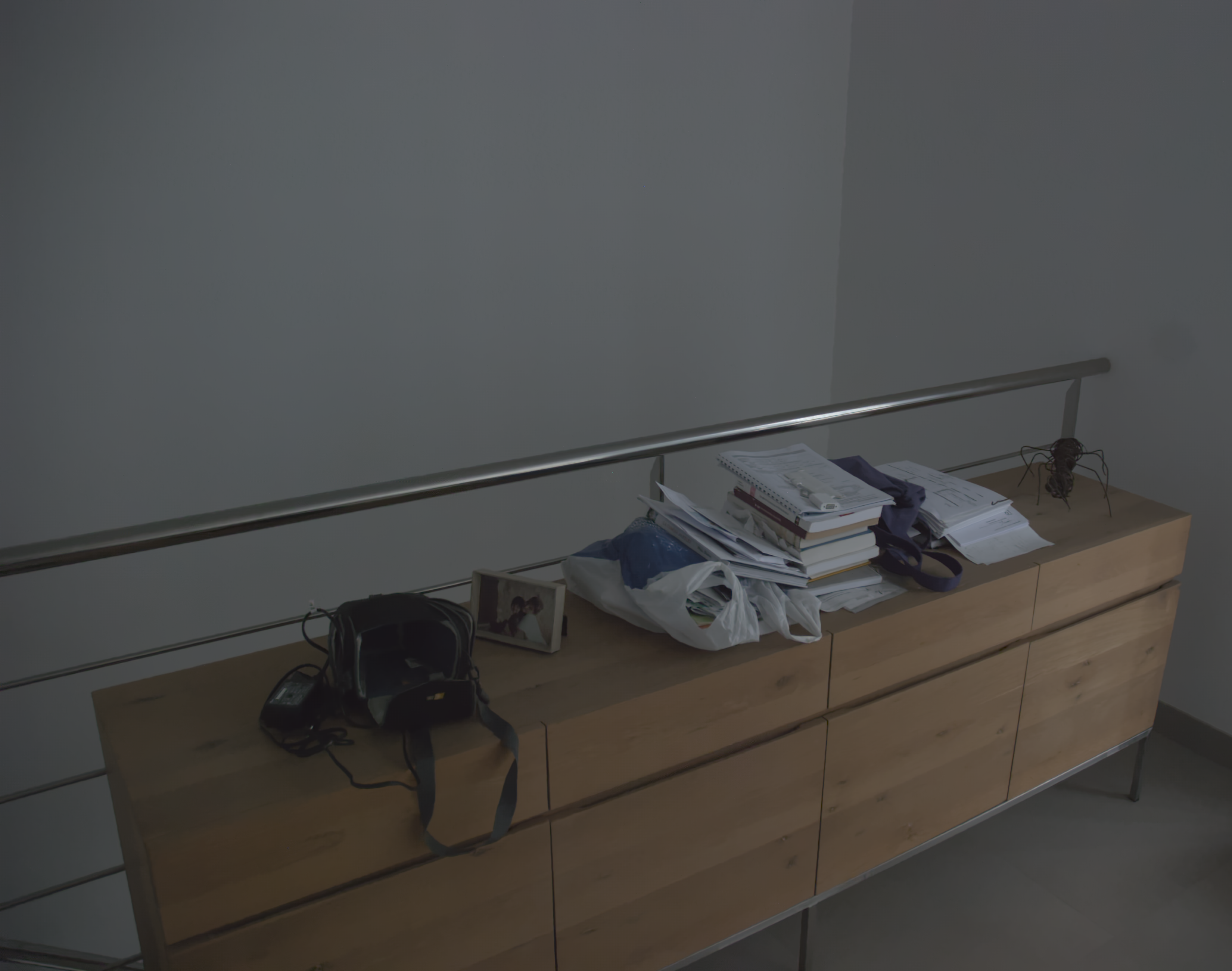}

\includegraphics[trim = 850px  650px 1800px 1550px, clip, width=5.2cm]{initial_crop.png}
\includegraphics[trim = 850px  650px 1800px 1550px, clip, width=5.2cm]{restoredTCSVTsinleIt.i1.png}
\includegraphics[trim = 850px  650px 1800px 1550px, clip, width=5.2cm]{denoised_fl0.85_f1.5_crop.png}

\includegraphics[trim = 2000px  950px 900px 1350px, clip, width=5.2cm]{initial_crop.png}
\includegraphics[trim = 2000px  950px 900px 1350px, clip, width=5.2cm]{restoredTCSVTsinleIt.i1.png}
\includegraphics[trim = 2000px  950px 900px 1350px, clip, width=5.2cm]{denoised_fl0.85_f1.5_crop.png}

\caption{Nikon general sequence. From left to right:  image formed by directly applying the imaging chain, the video denoising algorithm proposed in \cite{buades2017denoising} and the proposed chain.} \label{fig:nikon_general_sequence}
\end{figure*}

Figure \ref{fig:nikon_general_sequence} compares  the application of these methods to a sequence of images acquired with the Nikon D80.  Despite being an indoor sequence,   the scene is quite illuminated and the noise is moderate. The colored noise spots of the non denoised image are easily noticeable. The denoised sequence by using \cite{buades2017denoising}, having access only to the images after the imaging pipeline is applied, is not able to completely remove noise, and isolated color spots remain.  The proposed chain is able to completely remove noise while keeping the main details. 

\begin{figure*}[t]
\centering

\includegraphics[width=5.2cm]{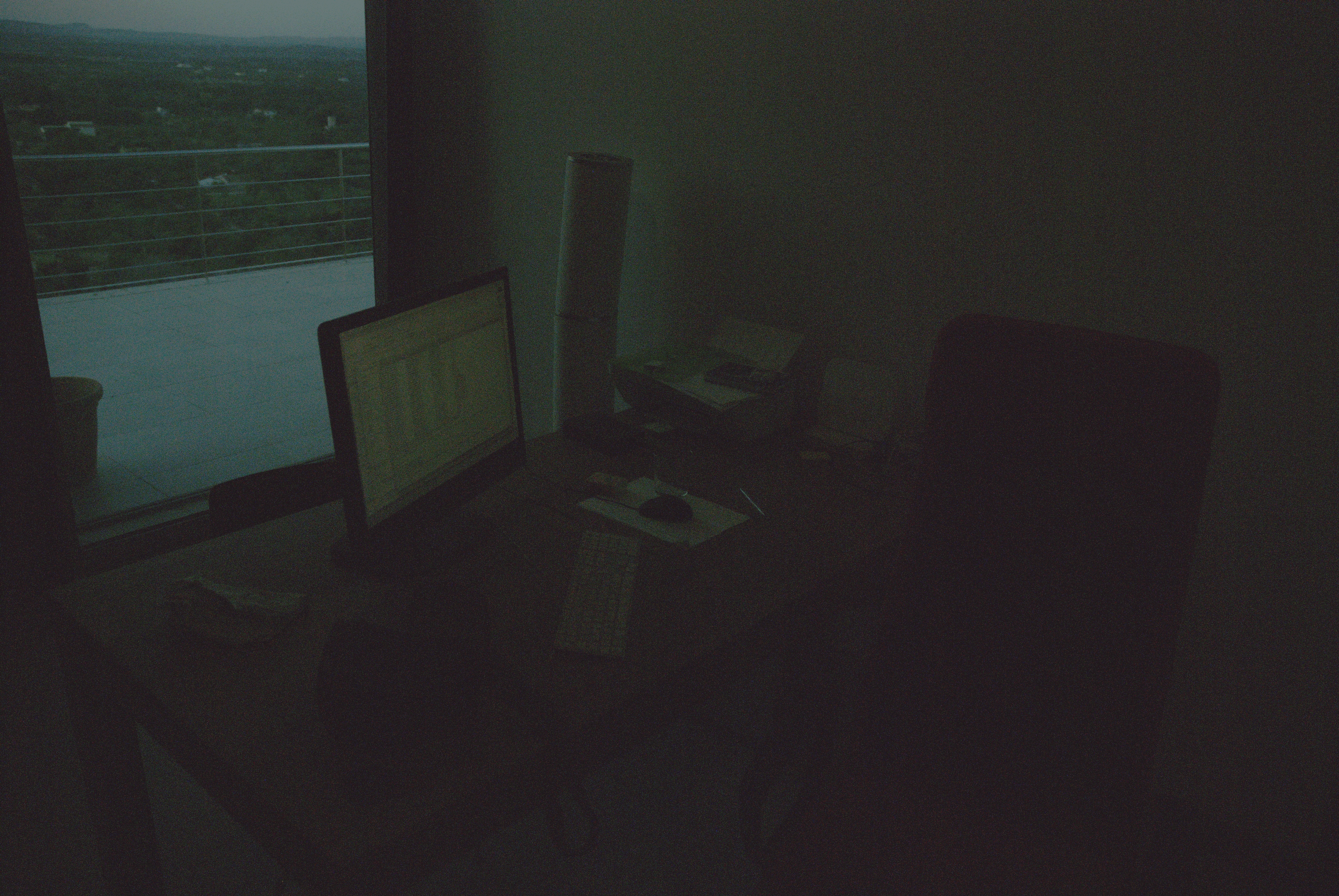}
\includegraphics[width=5.2cm]{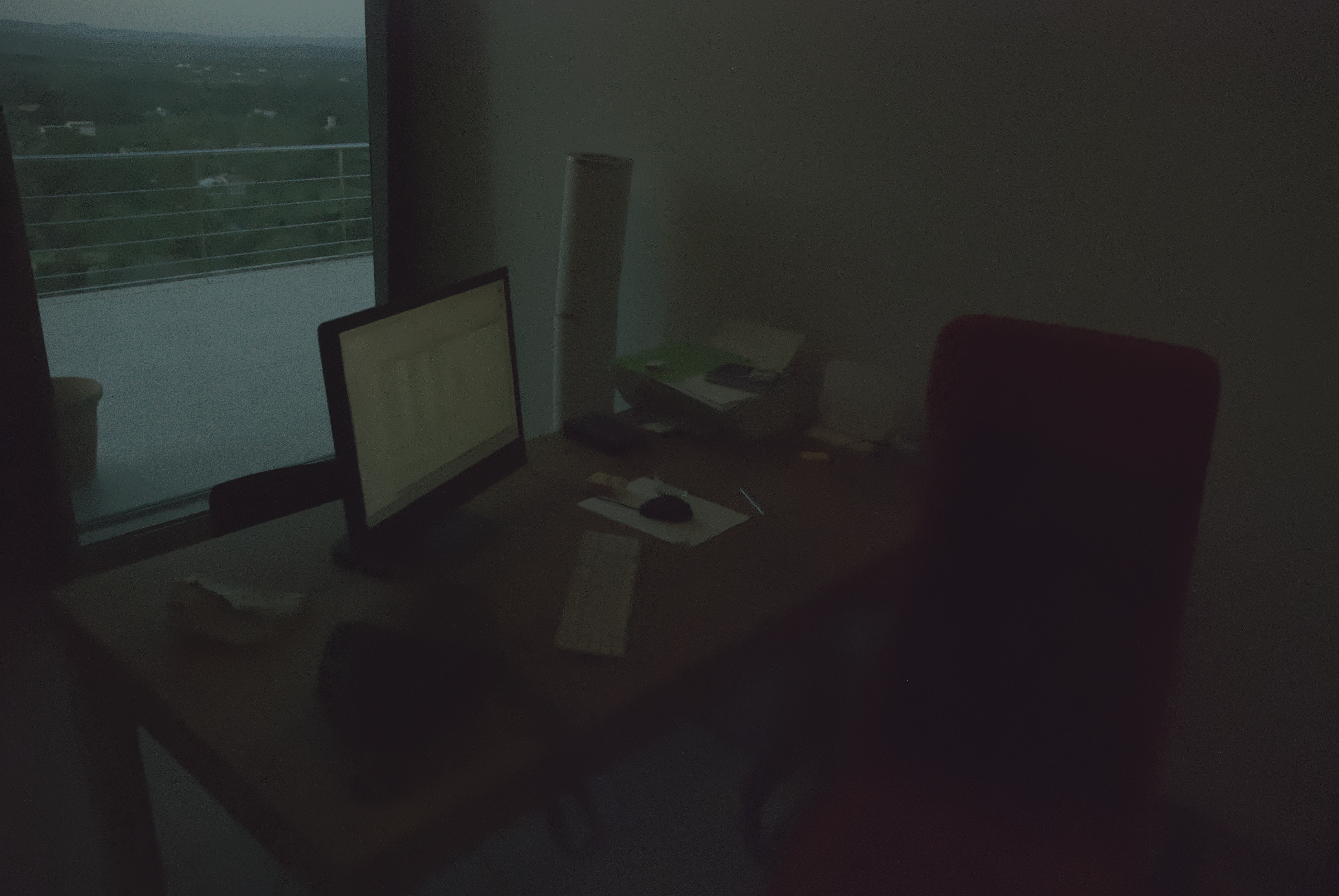}
\includegraphics[width=5.2cm]{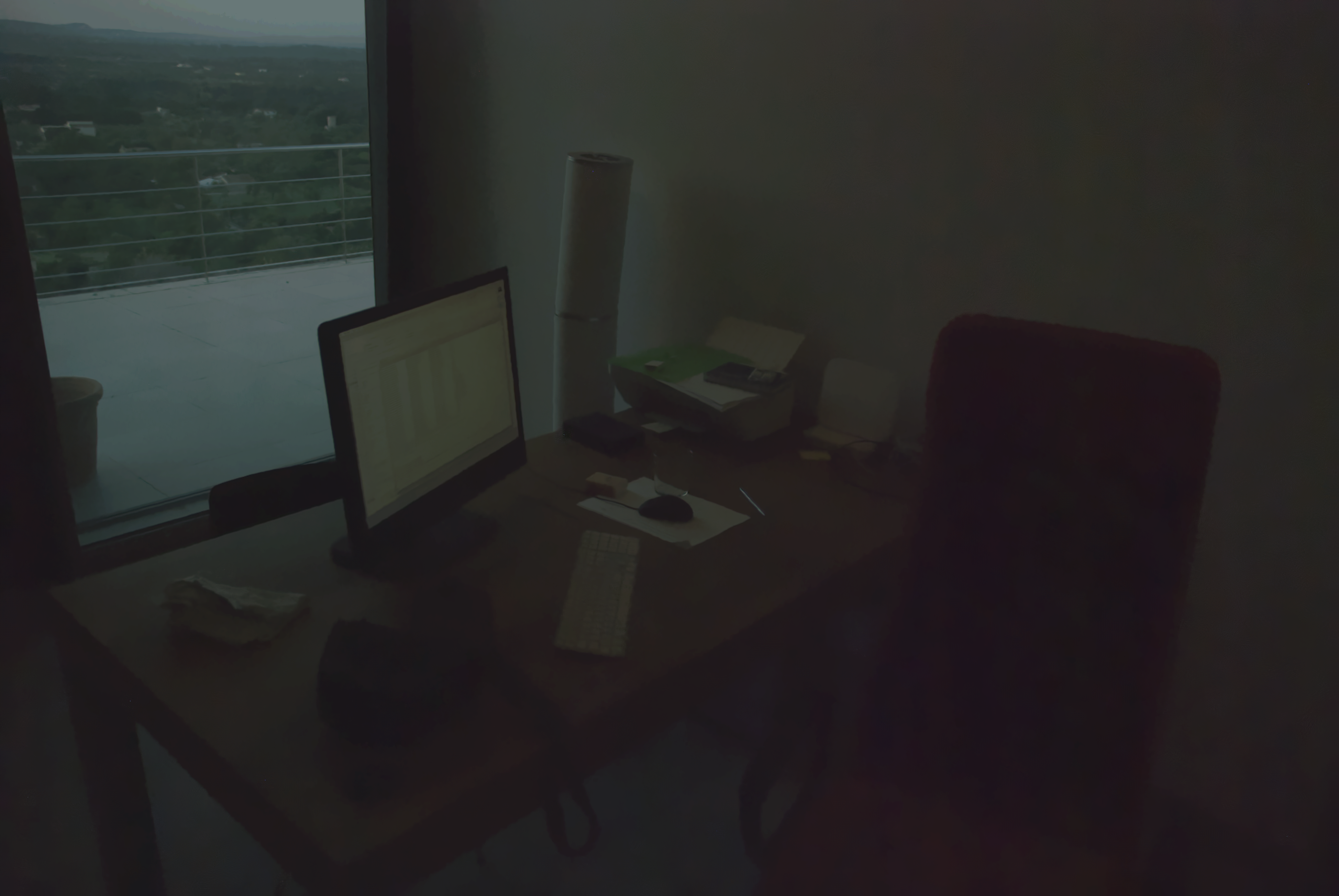}

\includegraphics[trim = 1350px  950px 1800px 1150px, clip, width=5.2cm]{hscinitial_crop.png}
\includegraphics[trim = 1350px  950px 1800px 1150px, clip, width=5.2cm]{hscrestoredTCSVTsinleIt.i1.png}
\includegraphics[trim = 1350px  950px 1800px 1150px, clip, width=5.2cm]{hscdenoised_fl0.85_f1.5_crop.png}

\includegraphics[trim = 200px  1650px 2700px 350px, clip, width=5.2cm]{hscinitial_crop.png}
\includegraphics[trim = 200px  1650px 2700px 350px, clip, width=5.2cm]{hscrestoredTCSVTsinleIt.i1.png}
\includegraphics[trim = 200px  1650px 2700px 350px, clip, width=5.2cm]{hscdenoised_fl0.85_f1.5_crop.png}

\caption{Nikon general sequence. From left to right:  image formed by directly applying the imaging chain, the video denoising algorithm proposed in \cite{buades2017denoising} and the proposed chain.} \label{fig:nikon_general_sequence_2}
\end{figure*}

Figure \ref{fig:nikon_general_sequence_2}  displays a similar experience but with darker conditions. The image with standard imaging chain illustrates the poor signal to noise ratio in these conditions, for which image details are hardly visible. 
The proposed algorithm is able to completely remove the noise, making many details appear despite they were hidden by noise. The method in \cite{buades2017denoising} is not able to correctly remove noise.

\subsection{Burst sequence of RAW images} 

We test the performance of the method with burst sequences and compare it with more adapted techniques for this type of data.
We took a burst of a poster pasted on a planar surface.  
It is well known, that in that case, a parametric tranformation is able to correctly register any two images of the sequence.  Such a parametric transformation should include radial distorsion parameters if the two view points are quite different.  Most methods estimate a global homography, a tiled translation, or even simpler transformations as a global affinity.

We compare our method with the bust method introduced in \cite{haro2012photographing}, which is composed by a global registration with an homography using SIFT \cite{Lowe04} characteristic points and a weighted combination of the registered images.
This burst method is applied to the images after the imaging chain is applied since we need to compute the characteristic points.  We also test the proposed method replacing the optical flow registration in the denoising and demosaicking stages  by a global homography registration using SIFT points.

\begin{figure*}[th]
\centering

\includegraphics[trim = 10px 10px 1400px 10px, clip, width=4cm]{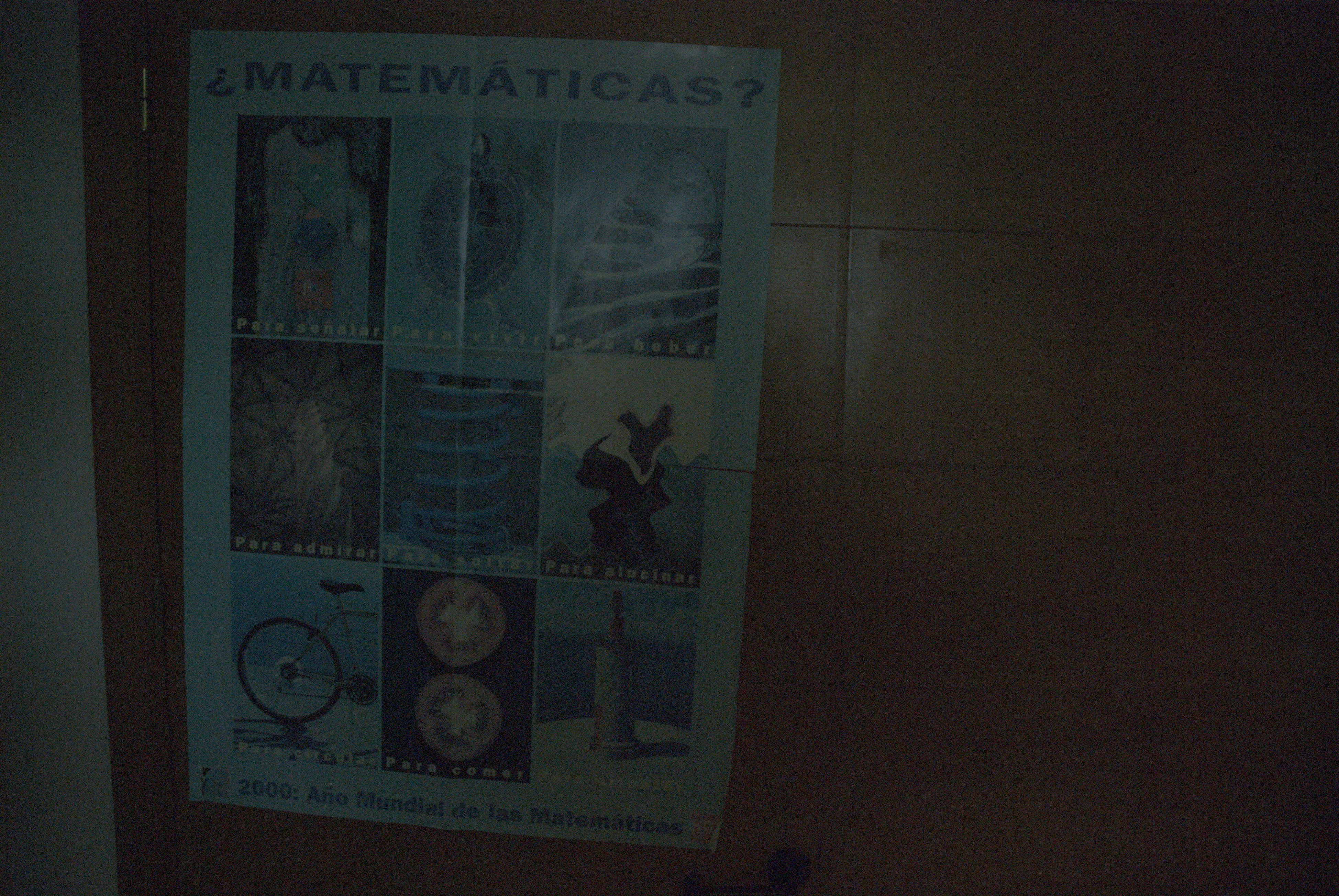}
\includegraphics[trim = 10px 10px 1400px 10px, clip, width=4cm]{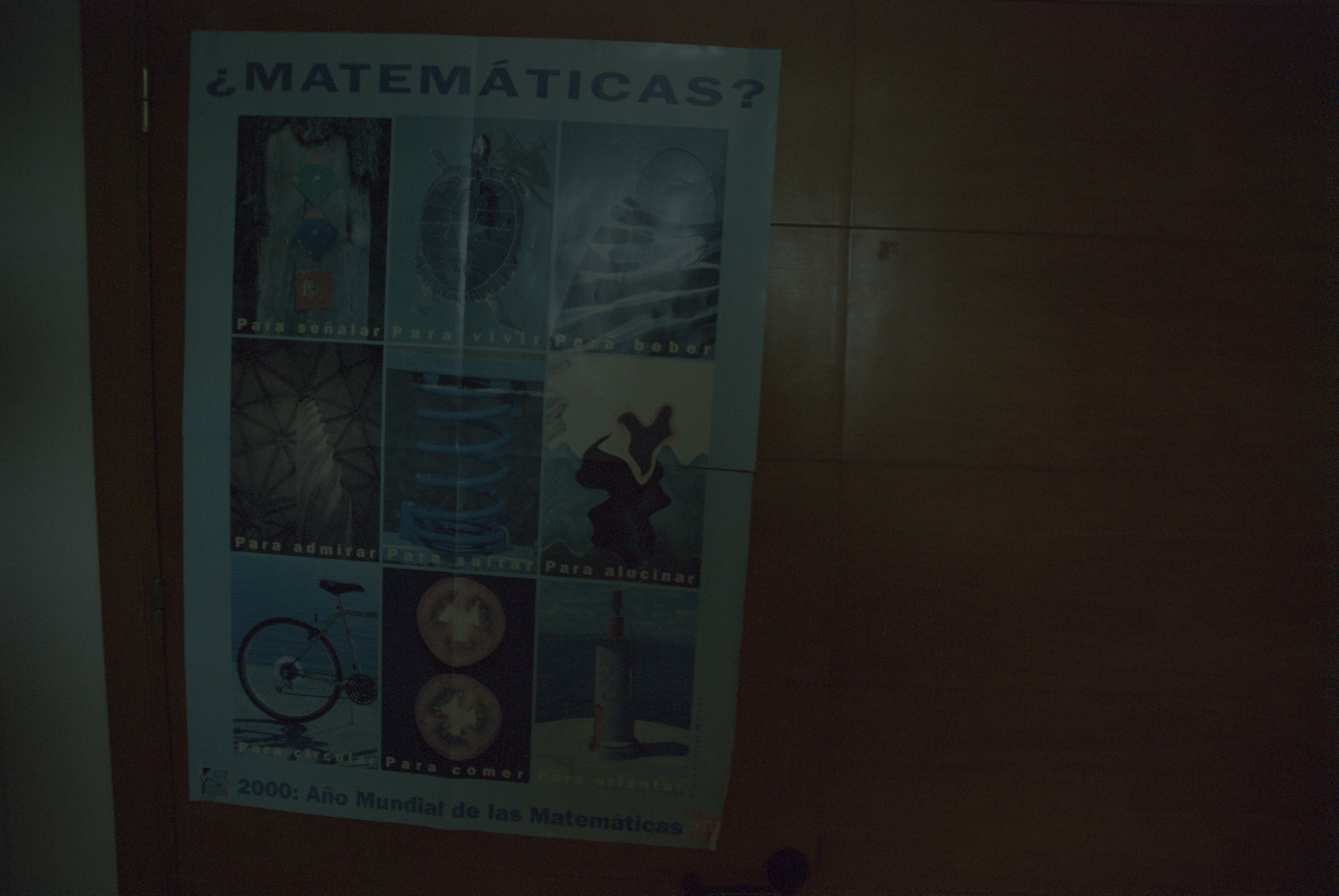}
\includegraphics[trim = 10px 10px 1400px 10px, clip, width=4cm]{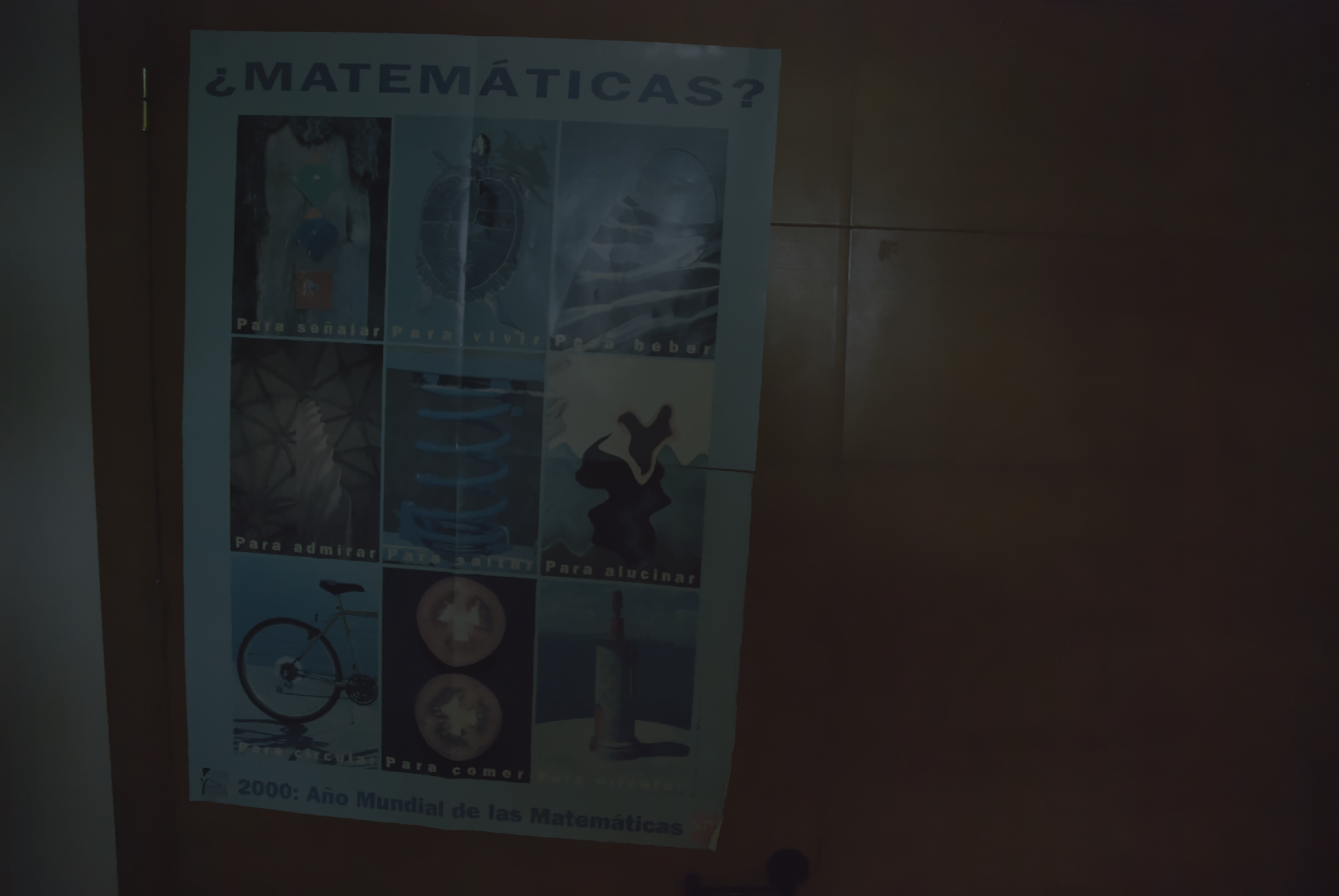}
\includegraphics[trim = 10px 10px 1400px 10px, clip, width=4cm]{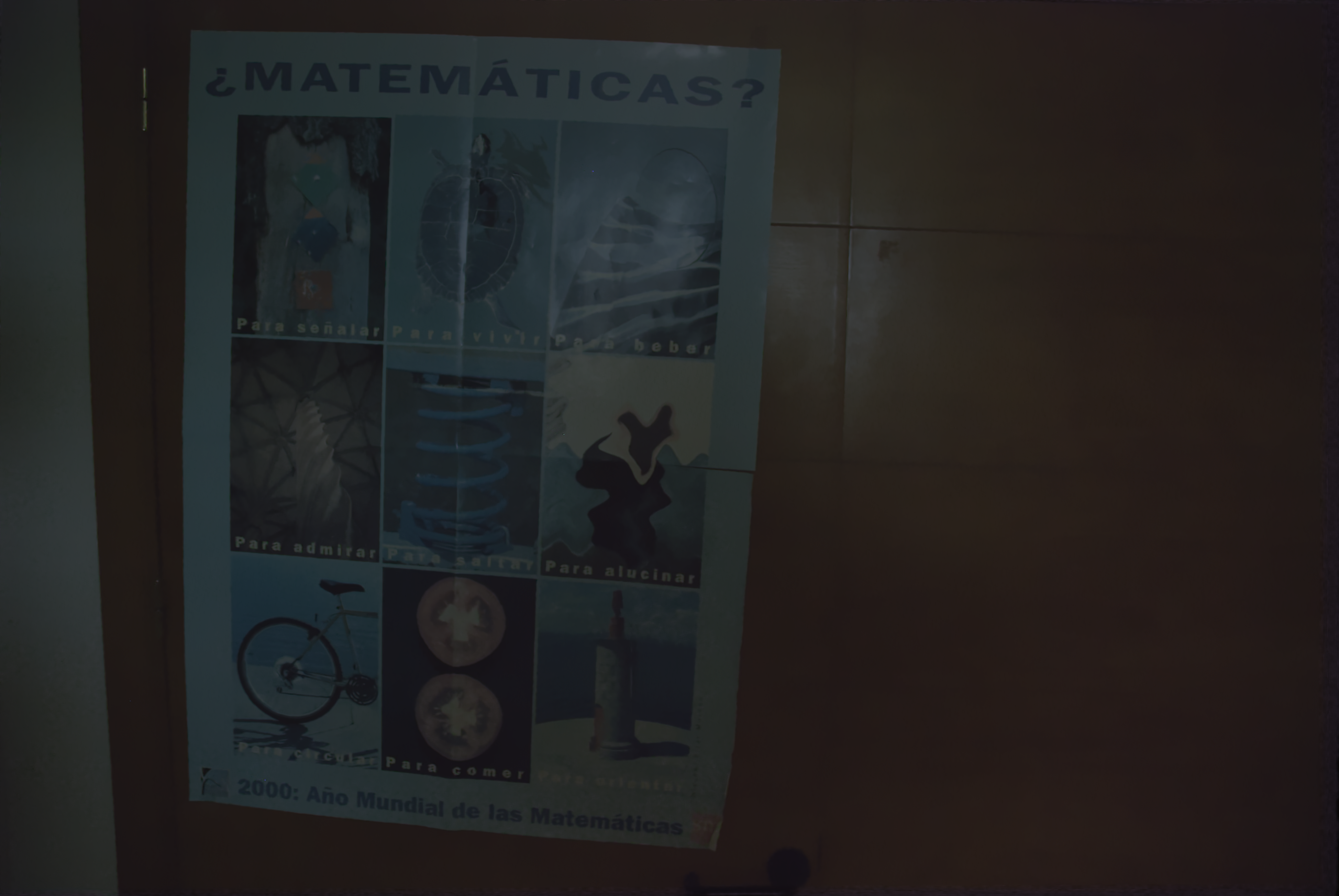}

\includegraphics[trim = 700px 1400px 2250px 300px, clip, width=4cm]{initial_crop10.png}
\includegraphics[trim = 700px 1400px 2250px 300px, clip, width=4cm]{burst_crop.png}
\includegraphics[trim = 700px 1400px 2250px 300px, clip, width=4cm]{denoised_fl0.85_f1.5_crop10.png}
\includegraphics[trim = 700px 1400px 2250px 300px, clip, width=4cm]{denoised_burst_raw_fl0.85_f1.5_crop.png}

\caption{Image burst sequence. From left to right:  image formed by directly applying the imaging chain,  the burst fusion \cite{haro2012photographing}, the proposed chain  and the proposed chain replacing all optical flow registration by a global homography estimation.} \label{fig:nikon_burst}
\end{figure*}

Figure \ref{fig:nikon_burst} displays the results. 
The burst algorithm is not able to remove completely noise, since its variance is reduced only as $1/M$ being $M$ the number of frames.  However, the denoised result is visually pleasant and has no artifacts. 
The proposed chain is able to recover all image details and completely removes noise.  Our result is not improved by using the global registration,  being the two solutions nearly identical. This shows the robustness of our method to the possible inaccuracies of the optical flow registration.

\section{Conclusions} \label{sec:conclusions}

We have introduced new denoising and demosaicking methods to deal with a sequence of RAW images. These methods make use of registration by optical flow and robust selection of similar patches by using spatio-temporal distances. The combination of both algorithms yields a sequence free of noise and demosaicking artifacts. 

The proposed strategy applies particularly well to image bursts. The comparison with state of the art methods illustrates the performance of the proposed method.

\bibliographystyle{plain}
\bibliography{references,references2}

\end{document}